\documentclass{ws-ijcis}

\usepackage[sort,compress]{cite}
\usepackage[verbose]{hyperref}
\usepackage{xcolor}
\hypersetup{colorlinks=false,allbordercolors=blue,pdfborderstyle={/S/U/W 1}}
\usepackage[linesnumbered, ruled, vlined]{algorithm2e}

\usepackage{graphicx}%
\usepackage{multirow}%

\usepackage{mathrsfs}%
\usepackage{listings}%

\usepackage{adjustbox} 
\usepackage{tabularx}
\usepackage{booktabs}
\usepackage[linesnumbered,ruled,vlined]{algorithm2e}
\usepackage{rotating}
\usepackage{textcomp}
\usepackage{subcaption}
\usepackage{caption}
\usepackage{lscape} 
\usepackage{booktabs} 
\usepackage{longtable} 
\usepackage{array}
\usepackage{longtable, array, booktabs}
\usepackage{wrapfig}  
\usepackage[table]{xcolor} 
\usepackage{xfp} 

\newcommand{\TemporalSweep}{\text{\(t_{sweep}\)}}

\newcommand{\traceprefix}{\text{\ensuremath{\mathcal{P}}}}

\newcommand{\activityinstancelog}{\ensuremath{\mathcal{L}}}

\newcommand{\activityinstancetraces}{\ensuremath{\sigma_c}}

\newcommand{\interstarttime}{\ensuremath{\delta^{\text{start}}_{c,k}}}
\newcommand{\interstarttimeplusone}{\ensuremath{\delta^{\text{start}}_{c,k+1}}}

\newcommand{\interprocessingtime}{\ensuremath{\delta^{\text{proc}}_{c,k}}}
\newcommand{\interprocessingtimeplusone}{\ensuremath{\delta^{\text{proc}}_{c,k+1}}}

\begin{document}

\markboth{Muhammad Awais Ali}{How Will My Business Process Unfold?}

\catchline{0}{0}{0000}{}{}

\title{How Will My Business Process Unfold? Predicting Case Suffixes With Start and End Timestamps}

\author{Muhammad Awais Ali\footnote{
Email: muhammad.awais.ali@ut.ee}}

\address{University of Tartu, Tartu, Estonia\footnote{ University of Tartu, Tartu, Estonia
}}

\author{Marlon Dumas}

\address{University of Tartu, Tartu, Estonia
}

\author{Fredrik Milani}

\address{University of Tartu, Tartu, Estonia
}

\maketitle


\begin{abstract}
Predictive process monitoring techniques support operational decision-making by predicting future states of ongoing cases in a business process. A subset of these techniques focuses on predicting the remaining sequence of activities for an ongoing case, known as case suffix prediction. Existing approaches for case suffix prediction typically generate sequences of activities with a single timestamp (e.g., the end timestamp). While this approach is useful in some contexts, it is insufficient for applications like resource capacity planning, where it is critical to reason about both the waiting time and processing time of each activity. Such information is necessary to accurately assess when resources will be engaged, optimize scheduling, and ensure capacity is adequate for future workloads. This paper introduces a technique for predicting case suffixes consisting of activities with both start and end timestamps. Specifically, the proposed technique predicts both the waiting time and processing time of each activity. Since the waiting time of an activity in a case depends on the availability of resources in other ongoing cases, the technique adopts a sweep-line approach, wherein the suffixes of all ongoing cases are predicted in lockstep, rather than predictions being made for each case in isolation. Additionally, the technique improves temporal accuracy by integrating resource availability information, automatically discovered from historical event logs, allowing for more realistic predictions in dynamic environments. An evaluation on real-life and synthetic datasets compares the accuracy of different instantiations of this approach, demonstrating the advantages of a multi-model approach to case suffix prediction over existing methods. The results highlight the improved performance in predicting both control-flow and temporal aspects of case suffixes, particularly in scenarios involving sparse and intermittent resource availability.
\end{abstract}

\keywords{Process Mining; Predictive Process Monitoring; Sequence Prediction}

\section{Introduction}

Predictive Process Monitoring (PPM) plays a role in Business Process Management (BPM) by providing managers with data-driven projections on running cases, enabling more informed and effective decision making~\cite{RamaManeiroVL23}. By leveraging models trained on event logs, PPM techniques offer runtime insights that predict the future states of ongoing process cases. For example, PPM techniques can predict the outcome of ongoing cases (e.g., will a customer accept or reject a product?)~\cite{TeinemaaDRM19}, the remaining time of ongoing cases~\cite{Verenich19}, the next activity in a case, or the sequence of remaining activities in a case (a.k.a. the \emph{case suffix})~\cite{RamaManeiroVL23}. These predictions are useful for process optimization, resource planning, and decision-making, especially in operational environments where timely insights are needed to, for instance, maintain efficiency and to meet service level agreements.

Existing approaches for predicting the next activities in a process, based on its current state (a.k.a. case suffix), primarily focus on generating sequences of activities with a single timestamp (e.g., end timestamp)~\cite{camargolstm,TaxVRD17,GunnarssonBW23}. This is insufficient for use cases such as capacity planning and scheduling, where managers need both start and end timestamps to calculate processing and waiting times, assess resource utilization, and determine whether current capacity is adequate for the expected workload~\cite{ali_data-driven_2024}. For instance, an operational manager may anticipate that, due to a flu epidemic, around 10\% of the workforce in a specific business area will report sick next Monday morning. As a result, reduced capacity to handle tasks is expected throughout the day. What is needed is the projected workload on Monday morning, given the current set of cases and their current state. If predictions only provide the next activities between now and the end of next week, using only one timestamp, it becomes impossible to determine how many activity instances will be active during specific periods, such as between 8 AM and 10 AM. Without the start timestamp, it is not possible to discern whether activity instances are in progress, waiting, or recently completed. In contrast, if predictions provide both start and end timestamps for each activity, the number of activity instances active between 8 AM and 10 AM on Monday can be estimated. This enables informed decisions about the number of workers needed to manage that workload effectively.

Previous approaches primarily focus on the temporal and control flow aspects of process prediction, such as the sequence of next activities and activity timestamps. However, they typically overlook resource availability and utilization. Including these factors can directly enhance insights by allowing a better understanding of how activity durations are affected by resource constraints, how resources should be allocated, and how many activities can be executed concurrently. For example, if resources are limited during a specific period, the predicted duration of an activity may increase, or certain activities might be delayed, which in turn impacts the overall schedule. This gap in current approaches highlights an opportunity for improvement. By incorporating both the start and end timestamps of predicted activities, along with resource availability and utilization, managers can gain a more comprehensive view of the process. Such predictions would enable them to make more informed decisions about resource allocation, ensuring that tasks are completed efficiently even during periods of resource constraints.

In this setting, we propose an approach for predicting case suffixes composed of activities with start and end timestamps. Our method uses a sweep-line-based technique~\cite{RafalinS08}, which incorporates resource availability features based on resource availability calendars, automatically discovered from historical event logs to predict case suffixes for all ongoing cases collectively, rather than individually. The approach follows a three-stage prediction process. First, a model predicts the next activity. Given this activity, a second model estimates the inter-start time, i.e., the time elapsed between the start of the previous activity and the start of the predicted activity. Finally, a third model predicts the processing time, i.e., the duration required to complete the activity once it has started. Using these predictions, we derive the start and end timestamps of each predicted activity instance in the case suffix. The start timestamp is obtained by adding the inter-start time to the start timestamp of the previous activity instance, while the end timestamp is derived by adding the processing time to the start timestamp of the predicted activity instance. This approach ensures accuracy by providing precise estimations of time intervals (inter-start time and processing time) that determine the timing of each activity. It is also complete in that it provides both start and end timestamps for each activity, offering a full timeline of the case suffix, which is essential for effective resource management, scheduling, and operational forecasting.


This article is an extended version of ~\cite{ali2025predicting}. The conference paper focuses on predicting case suffixes consisting of activities with start and end timestamps by employing a sweep-line-based approach. This article extends the conference paper by improving the temporal accuracy of the proposed approach by integrating information about resource availability based on resource availability calendars, automatically discovered from historical event logs. 
We hypothesize that this resource availability information, automatically discovered from historical event logs, will allow us to further refine our predictions and make them more robust and applicable to real-world scenarios, where resource constraints and availability influence process outcomes.


We conducted a two-pronged evaluation using synthetic and real-life event logs to assess the effectiveness of our multi-model approach. The synthetic evaluation examines how varying resource availability calendars affect the control-flow and temporal accuracy of case suffix prediction, simulating different on-off duty schedules, and resource constraints. This allows us to explore how the proposed approach adapts to changes in resource availability, which is a common challenge in real-world process management. Meanwhile, the evaluation with real-life logs compares our approach to existing multi-model approaches, using control-flow and temporal metrics to assess the accuracy and reliability of case suffix predictions in terms of both start and end timestamps.

The remainder of this article is structured as follows. Sec.~\ref{background} provides an overview of prior research related to case suffix prediction. Sec.~\ref{sec3} presents the proposed approach for case suffix prediction. Finally, Sec.~\ref{sec4} reports the empirical evaluation, and Sec.~\ref{conclusion} concludes the paper and Sec.~\ref{FW} outlines directions for future work.

\section{Related Work}\label{background}

Predictive Process Monitoring (PPM) is a research stream within the broader field of Process Mining (PM)~\cite{DumasRMR18}. Unlike traditional descriptive approaches such as process discovery, conformance checking, and model enhancement,  PPM incorporates predictive and monitoring capabilities that apply machine learning models to ongoing cases in a business process~\cite{VerenichDRMT19,TeinemaaDRM19}. By exploiting historical event data generated by organizational information systems, PPM transforms process analysis from a backward-looking to a forward-looking perspective~\cite{Francescomarino17}. Through this predictive capability, it becomes possible to estimate future activities, remaining execution time, or likely case outcomes, which are essential for data driven decision making~\cite{di2018predictive,Francescomarino17}. In practice, predictive models derived from PPM can support the timely allocation of resources~\cite{MARTIN2020101463}, early identification of process bottlenecks~\cite{ToosinezhadFKA20}, and adaptive interventions aimed at improving operational performance~\cite{DavidDumas,ShoushD23}.

In recent years, PPM research has increasingly focused on employing deep learning architectures to enhance the predictive performance of forecasting future activities within an ongoing case (a.k.a the case suffix prediction). Early studies such as those by Tax et al.~\cite{Taxsuffixprediction} and Evermann et al.~\cite{EvermannRF17} introduced LSTM-based models to predict the next activity, case suffix, and remaining time until completion, using one-hot and embedded sequence encodings. Camargo et al.~\cite{camargolstm} extended these efforts through multi-output LSTM models that jointly predict activity, end timestamp, and role. Other approaches have leveraged architectures, such as GANs~\cite{TaymouriREBV20}, transformers and attention-based encoder-decoder models~\cite{WuytsBW24, RamaManeiroVLM24}, and reinforcement learning-based samplers like DOGE~\cite{RamaManeiroPVL24}. Pasquadibisceglie et.al.,~\cite{Pasquadibisceglie19}  also explored CNNs and, more recently, large language models (LLMs)~\cite{PasquadibisceglieAM24} for case suffix prediction. 

The above approaches follow a single model (SM) paradigm, which jointly predicts the next activity and one timestamp (either the start or the end time). None of these approaches is designed to estimate both the start and the end timestamps of the activities in the case suffix. In other words, none of these approaches separates the waiting time and the processing time of the activities in the suffix. Our approach tackles this limitation by departing from the SM approach and, instead, training three separate models individually to predict the next activity, its inter-start time, and its processing time. This approach allows us to flexibly calculate waiting times and processing times separately.

Furthermore, researchers have explored the integration of learning-based predictors with simulation models. Camargo et al.~\cite{CamargoBDR23} and Meneghello et al.~\cite{MeneghelloFGR25} combined LSTM predictors with discrete-event simulation derived from discovered process models to generate synthetic traces and evaluate “what-if” scenarios. These hybrid methods are valuable for studying process changes or capacity adjustments. In contrast, our approach focuses on generating case suffixes for ``as-is" processes by leveraging real-time event logs without assuming process changes or relying on synthetic generation.

Recently, a few of the studies have focused on integrating resource-related features to enhance both predictive accuracy and model interpretability~\cite{KlijnTFM24,KunklerR24,KirchdorferBKAS24,TourPKS23}. Klijn et al.~\cite{KlijnTFM24} investigated resource waiting-time patterns to better understand process performance, showing that analyzing behavior at the resource level yields more meaningful and interpretable insights than relying on aggregated waiting-time averages. Similarly, Kunkler et al.~\cite{KunklerR24} demonstrated that considering the time individual resources require to complete tasks can outperform conventional allocation strategies, underscoring the role of human and resource characteristics in process optimization. Additionally, a recent survey by Ali et al.~\cite{ali_data-driven_2024} explored the various causes and interpretations of waiting times related to resource contention and unavailability, aiming to better understand process bottlenecks through the analysis of resource behavior.

Beyond predictive modeling, several studies have revisited business process simulation (BPS) from a resource-oriented standpoint. For example, the authors in~\cite{KirchdorferBKAS24} proposed a resource-first simulation approach that constructs multi-agent systems from event logs, effectively capturing heterogeneous resource behaviors and interactions. Their findings indicated improved accuracy, reduced computational overhead, and greater interpretability across various process settings. In the same direction, Tour et al.~\cite{TourPKS23} introduced Agent Miner, an algorithm designed to discover agent interaction models directly from event data, resulting in representations of business processes that more closely mirror real-world execution compared to traditional discovery techniques.

Despite recent advancements in case suffix prediction using PPM, most approaches remain focused primarily on control-flow aspects~\cite{KratschMRS21}. A recent benchmark and review by Rama et al.~\cite{RamaManeiroVL23} highlights that newer techniques, such as LSTMs and other deep learning models, continue to adopt a case-specific control-flow perspective, often overlooking the influence of resource availability factors. While previous studies have incorporated resource-related information, such as the role or resource performing an activity, no study to date has systematically integrated resource availability aware features such as working shifts, scheduled breaks, workload conditions, or expected availability times into predictive models for the case suffix prediction problem. Collectively, these gaps suggest that integrating resource availability aware information could enhance both the accuracy and performance of process mining tasks. Building on this foundation, our study advances the PPM literature by approaching the case suffix prediction problem from a resource-centric perspective, bridging predictive performance with the analysis of resource availability aware behavior for more accurate predictions of case suffixes in an ongoing business process.

\section{Approach}~\label{sec3}
The proposed approach consists of two phases: an offline phase for training predictive models for case suffix prediction and an online phase which adopts a sweep-line based method wherein the suffixes of all ongoing cases in the process are predicted in lockstep, rather than predictions being made in isolation. 

\subsection{Offline Phase}\label{offlinephase}

\begin{figure}[hbtp]
\includegraphics[width=\linewidth]{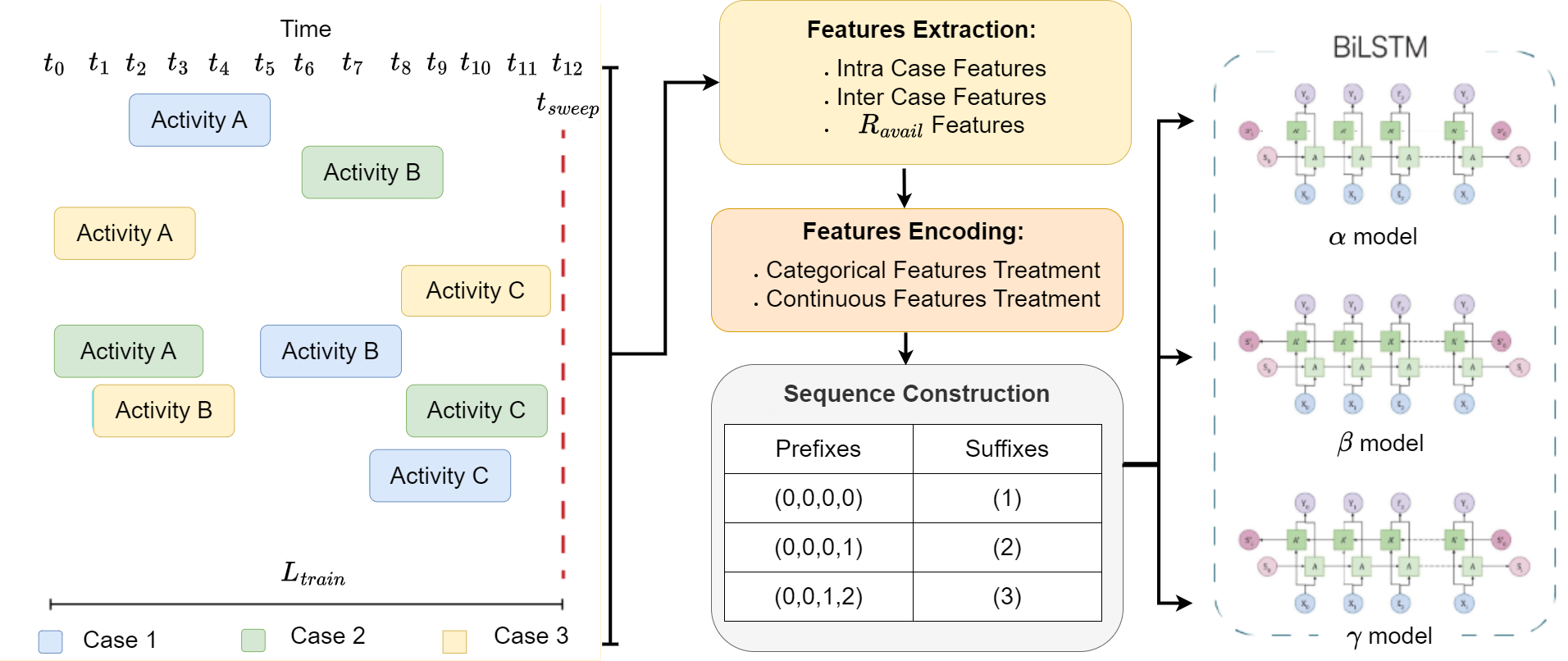}
   \caption{Offline Phase}
    \label{offlinefig}
\end{figure}

The offline phase (Fig.\ref{offlinefig}) focuses on preparing the training data and learning predictive models necessary for case suffix prediction. This phase begins with transforming an event log into structured input sequences by extracting intra-case features (capturing case-specific temporal patterns), resource contention features (capturing system-wide dynamics, such as resource utilization and workload) and resource availability features. These features are encoded and assembled into fixed-size sequences using an n-gram strategy to standardize inputs for model training. We train three specialized BiLSTM-based models -- one each for predicting the next activity($\alpha$), inter-start time($\beta$), and processing time($\gamma$). Below, we describe the offline phase.\medskip

\subsubsection{Input} To train models for case suffix prediction, we take as input a type of event log called an \emph{activity instance log}.  Table~\ref{instace_log} contains an excerpt of such a log. Each row corresponds to an activity instance. For each activity instance, the proposed approach requires a case ID, activity name, a resource, start and end timestamps. Every row must have a value for each of these five attributes. However, for some activity instances (e.g., the second one in Case 2) the value of the end timestamp may be null (denoted as $\emptyset$). When an activity instance has a null end timestamp, it means that it has not yet completed.  On the other hand, the approach does assume that the case ID, activity name, start and end timestamps are accurate, otherwise the machine learning models would be trained to produce inaccurate case suffixes. Formally, we define an activity instance log as follows.



\begin{definition}[\textbf{Activity Instance Log}] 
An \emph{Activity Instance Log} \activityinstancelog, is a finite set of records \( e \in \activityinstancelog\), where each record \( e = (c, a, r, T^{\text{start}}, T^{\text{end}}) \) consists of the following elements: \( c \in C \), where \( C \) is the set of unique \emph{case identifiers} for process instances; \( a \in A \), where \( A \) is the set of possible \emph{activities}; \( r \in R \), where \( R \) is the set of \emph{resources} executing the activity; and \( T^{\text{start}}, T^{\text{end}} \in \mathbb{T} \cup \{\emptyset\} \), representing the \emph{start timestamp} and \emph{end timestamp} of an activity, with \( \mathbb{T} \) being the domain of time values and $\emptyset$ denotes a missing (null) timestamp.
\end{definition}

\begin{table}[hbtp]
\centering
\footnotesize
\caption{Activity Instance Log}

\setlength{\tabcolsep}{10pt}   

\begin{tabular}{@{}c  l l c c@{}}
        \toprule
        \textbf{Case ID} & \textbf{Activity } & \textbf{Resource} & \textbf{$T^{\text{start}}$} & \textbf{$T^{\text{end}}$} \\
        \midrule
        \multirow{4}{*}{1} & Received Query          & System        & 08:00 & 08:05 \\
                           & Assigned to Rep         & Supervisor    & 08:10 & 08:15 \\
                           & Query Resolved          & Rep~A         & 08:20 & 09:00 \\
                           & Cust Notified           & System        & 09:05 & 09:10 \\
        \midrule
        \multirow{2}{*}{2} & Received Query          & System        & 09:15 & 09:20 \\
                           & Assigned to Rep         & Supervisor    & 09:25 & \(\emptyset\) \\
        \midrule
        \multirow{3}{*}{3} & Received Query          & System        & 10:00 & 10:05 \\
                           & Assigned to Rep         & Supervisor    & 10:10 & 10:15 \\
                           & Query Resolved          & Rep~B         & 10:20 & 11:00 \\
        \bottomrule\\
    \end{tabular}

\label{instace_log}
\end{table}



The Activity Instance Log (\activityinstancelog{}) may be structured as a set of activity instance traces (\activityinstancetraces), where each trace is an ordered sequence of activities for a specific case identifier (\( c \)). An activity instance trace is defined as follows.

\begin{definition}[\textbf{Activity Instance Trace of Log \activityinstancelog{}}]
\label{def2}
Given an activity instance log \activityinstancelog{}, an \emph{activity instance trace} \activityinstancetraces{} of \activityinstancelog{} is a sequence of activity instances \( \sigma_c = \langle e_{c,1}, e_{c,2}, \dots, e_{c,n} \rangle \), such that:
\begin{itemize}
    \item All activity instances $e_{c,i}$ in $\sigma_c$ share the same case identifier $c$. 
    \item Every activity instance in $\activityinstancelog{}$ that has $c$ as its case identifier is part of $\sigma_c$.
    \item The activity instances in $\sigma_c$ are chronologically ordered by start timestamp.
\end{itemize}
Herein, we write $\sigma_c \in \activityinstancelog{}$ to denote that $\sigma_c$ is a trace of log $\activityinstancelog{}$.
\end{definition}

Our approach takes as input an activity instance log  \activityinstancelog{} generated by a collection of cases over a specified time period and a time point called  \textit{cutoff} denoted by ($t_{\text{cutoff}}$). Given \(\activityinstancelog{}\) and $t_{\text{cutoff}}$, we derive the \emph{training log}, denoted as \(\activityinstancelog_{train}\). The training log includes all \(\activityinstancetraces{} \in \activityinstancelog{}\) that fulfil the following two conditions:

\begin{itemize}
\item Every activity instance in \(\activityinstancetraces{}\) has a non-null end timestamp, i.e., $\neq$ \(\emptyset\).
\item Every activity instance in \(\activityinstancetraces{}\) has an end timestamp $\leq$ $t_{\text{cutoff}}$.
\end{itemize}

To train a deep learning model, we enhance each case in the training log with resource contention, intra-case and resource availability features. Alg.~\ref{alg:enhance_traceprefix} describes the procedure for enhancing a given trace \(\activityinstancetraces{}\) with such features. This procedure is applied to each $\activityinstancetraces{}$  $\in$ $\activityinstancelog_{train}$.

\begin{algorithm}[htbp]
\caption{Enhancing an activity instance trace $\sigma_c$}
\label{alg:enhance_traceprefix}
\KwIn{An activity instance log $L$, an activity instance trace $\sigma_c \in L$, resource calendars $C_r$ for all resources}
\KwOut{An enhanced activity instance trace $\sigma_c$ including intra-, $R_{\text{cont}}$, and $R_{\text{avail}}$ features}

\ForEach{trace prefix $P = \langle e_{c,1}, \dots, e_{c,k} \rangle$ in $\sigma_c$}{
    
    \For{$i \leftarrow 1$ \KwTo $k$}{
        Let $e_{c,i} = (c, a_{c,i}, T^{\text{start}}_{c,i}, T^{\text{end}}_{c,i})$\;
        $\delta^{\text{proc}}_{c,i} \leftarrow T^{\text{end}}_{c,i} - T^{\text{start}}_{c,i}$\;
        \If{$i = 1$}{
            $\delta^{\text{start}}_{c,1} \leftarrow 0$\;
        }
        \Else{
            $\delta^{\text{start}}_{c,i} \leftarrow T^{\text{start}}_{c,i} - T^{\text{start}}_{c,i-1}$\;
        }
        $e_{c,i} \leftarrow e_{c,i} \oplus (\delta^{\text{proc}}_{c,i}, \delta^{\text{start}}_{c,i})$\;
    }

    Let $T^{\text{current}} \leftarrow T^{\text{start}}_{c,k}$\;

    \texttt{BusyResources} $\leftarrow 0$\;
    \texttt{TotalActiveInstances} $\leftarrow 0$\;

    \ForEach{$e = (c', a', T'_{\text{start}}, T'_{\text{end}}) \in L$}{
        \If{$T'_{\text{start}} \leq T^{\text{current}} \leq T'_{\text{end}}$}{
            \texttt{TotalActiveInstances} $\leftarrow$ \texttt{TotalActiveInstances} $+ 1$\;
            \If{$a' = a_{c,k}$}{
            \texttt{BusyResources} $\leftarrow$ \texttt{BusyResources} $+ 1$\;
            }
        }
    }

    $\text{WIP}_{c,k} \leftarrow$ \texttt{TotalActiveInstances}\;
    $e_{c,k} \leftarrow e_{c,k} \oplus (\text{WIP}_{c,k})$\;

    $\text{Utilization}_{c,k} \leftarrow |\texttt{BusyResources}|$\;
    $e_{c,k} \leftarrow e_{c,k} \oplus (\text{Utilization}_{c,k})$\;

    Let $\tau \leftarrow 0.2 \times \text{log duration}$\;
    \texttt{RecentCaseStarts} $\leftarrow \emptyset$\;

    \ForEach{case $c' \in L$}{
        Let $T^{\text{start}}_{c'}$ be the first start timestamp in $\sigma_{c'}$\;
        \If{$T^{\text{start}}_{c'} \geq T^{\text{current}} - \tau$}{
            Add $c'$ to \texttt{RecentCaseStarts}\;
        }
    }

    $\lambda_{c,k} \leftarrow |\texttt{RecentCaseStarts}| / \tau$\;
    $e_{c,k} \leftarrow e_{c,k} \oplus (\lambda_{c,k})$\;

   \tcp{--- Resource Availability Features ---}

$\mathcal{R}^{a_k} \leftarrow 
\mathcal{R}( a_{c,k})$\;

\ForEach{$r \in \mathcal{R}^{a_k}$}{
    $T^{r}_{\text{next\_on\_duty}} \leftarrow 
    \textsc{NextOnDuty}(C_r, T^{\text{current}})$\;

    $T^{r}_{\text{since\_last\_on\_duty}} \leftarrow 
    \textsc{SinceLastOnDuty}(C_r, T^{\text{current}})$\;

    $T^{r}_{\text{next\_off\_duty}} \leftarrow 
    \textsc{NextOffDuty}(C_r, T^{\text{current}})$\;

    $T^{r}_{\text{since\_last\_off\_duty}} \leftarrow 
    \textsc{SinceLastOffDuty}(C_r, T^{\text{current}})$\;
}

\ForEach{$f \in \{
T_{\text{next\_on\_duty}},
T_{\text{since\_last\_on\_duty}},
T_{\text{next\_off\_duty}},
T_{\text{since\_last\_off\_duty}}
\}$}{
    $\bar{f}(T^{\text{current}}) \leftarrow
    \min_{r \in \mathcal{R}^{a_k}} f(r, T^{\text{current}})$\;
}

$\bar{R}_{\text{avail}} \leftarrow
(\bar{f}_1, \bar{f}_2, \bar{f}_3, \bar{f}_4)$\;

$e_{c,k} \leftarrow e_{c,k} \oplus (\bar{R}_{\text{avail}})$\;

}
\end{algorithm}

\subsubsection{Basic (Intra-Case) Features}
We extract features from each \emph{trace prefix} of a trace $\sigma_c$. Given a trace 
\(\sigma_c = \langle e_{c,1}, \dots, e_{c,n} \rangle \) capturing the execution of case $c$, and given an index \( 1 \leq k \leq n \), the trace prefix of case $c$ at index $k$, denoted $\traceprefix(c, k)$, is the sequence \( \langle e_{c,1}, \dots, e_{c,k} \rangle \).  For example, in Table~\ref{instace_log}, the prefix of case 1 at index 2 
\( \traceprefix(1,2) = \langle (\text{Received Query}, \text{System}, 08{:}00, 08{:}05), (\text{Assigned to Rep}, \text{Supervisor}, 08{:}10, 08{:}15) \rangle \).


For each activity instance \( e_{c,k} \) = \(
(c, a_{c,k}, r_{c,k}, T^{\text{start}}_{c,k}, T^{\text{end}}_{c,k})
\) in a prefix, we construct an enhanced  activity instance \( (c, a_{c,k}, r_{c,k}, T^{\text{start}}_{c,k}, T^{\text{end}}_{c,k}, \delta^{\text{start}}_{c,k}, \delta^{\text{proc}}_{c,k}) \), where \( \delta^{\text{start}}_{c,k} = T^{\text{start}}_{c,k} - T^{\text{start}}_{c,k-1} \) denotes the inter-start time, and \( \delta^{\text{proc}}_{c,k} = T^{\text{end}}_{c,k} - T^{\text{start}}_{c,k} \) is the processing time. For the first activity, we define \( \delta^{\text{start}}_{c,1} = 0 \). The procedure for extracting vector $v$ is implemented in lines (\texttt{1-8}) of Alg.~\ref{alg:enhance_traceprefix}. The algorithm iterates over each trace \( \sigma_c \in \mathcal{L} \), where \( \mathcal{L} \) is the activity instance log grouped by case. For each trace prefix \( \traceprefix \subseteq \sigma_c \), it processes every activity instance, computes the associated intra-case features, and concatenates these features with the tuple representing activity instance \( \alpha_{c,i} \) with these features (line \texttt{9}).\footnote{Operator \( \oplus \) denotes tuple concatenation.}

\subsubsection{Resource Contention Features}

On top of the above basic features, we add to the feature vectors a set of features to capture resource contention, i.e.\ the level of busyness of resources. Resource contention is a determinant of waiting times in resource-constrained systems. Accordingly, we expect that these features will enhance the accuracy of inter-start-time predictors.

Prior research~\cite{GunnarssonBW24} highlights the importance of including \emph{work in progress (WIP)} and process load (e.g., resource utilization) in the context of next-activity and remaining time prediction. Unlike the basic features introduced above, which are extracted from a single trace at a time (intra-case calculation), the WIP is a resource contention feature. Its calculation requires reasoning across multiple cases.

To capture resource contention we propose to capture resource contention by means of three features: \emph{WIP}, resource utilization, and the arrival rate of new cases. The reason for including  \emph{WIP} and the arrival rate is because of the Little’s Law~\cite{DumasRMR18}. According to Little’s Law, the average cycle time (\(T_c\)) of a process is determined by the \textit{Work in Progress (WIP)} and arrival rate (\(\lambda\)) via \(WIP = \lambda \times T_c\). In human-centric processes, cycle time is often dominated by waiting time. We, therefore, hypothesize that features correlated with cycle time can improve waiting time prediction. Queuing theory suggests that waiting times are also influenced by \emph{resource utilization}, the ratio of resource demand to availability~\cite{ali_data-driven_2024}. When demand exceeds capacity, utilization reaches 100\%, causing delays, whereas Lower demand reduces utilization. Since our objective is not to predict resources, the task is simplified to predicting the activity instance within a case suffix. Accordingly, we leverage all available and relevant information to effectively predict the case suffix without incorporating resource attributes from the activity instance log.

Alg.~\ref{alg:enhance_traceprefix} explains the extraction of resource contention features,  denoted as ${R}_{\text{cont}}$, for a given \traceprefix.  It computes three resource contention features: \emph{WIP}, resource utilization, and the arrival rate of new cases (lines \texttt{10-29}). To compute WIP\footnote{Throughout this paper, WIP refers to the number of cases currently in execution at a given point in time, i.e., cases that are actively being processed.} at time \(T^{\text{start}}_{c,k}\), Alg.~\ref{alg:enhance_traceprefix} counts active activity instances in \(\activityinstancelog{}\) (lines \texttt{11-17}). It also estimates the arrival rate as \(\lambda = \frac{\text{Total cases}}{\text{Time window}}\) (lines \texttt{22-28}), using a recent time window (default: 20\% of log duration). The set of recent cases is updated dynamically (line \texttt{27}). Higher WIP and \(\lambda\) indicate system congestion, leading to longer expected waiting times. Alg.~\ref{alg:enhance_traceprefix} computes utilization by counting active instances per activity at a given time (lines \texttt{16-20}), helping predictive models account for resource load and capacity. For each trace prefix \( \traceprefix \subseteq \sigma_c \), Alg.~\ref{alg:enhance_traceprefix} computes resource contention features and augments \( e_{c,k} \) accordingly. In addition, Algorithm~\ref{alg:enhance_traceprefix} computes resource availability features using precomputed resource availability calendars, capturing temporal aspects of resource availability; these features are described and discussed in detail in the following section.

\subsubsection{Resource Availability Features}

In the context of business process simulation, calendars are used to model the availability of resources~\cite{KlijnTFM24,KunklerR24}. A resource availability calendar consists of a set of time intervals during which a resource is on-duty, i.e.\ available to perform activity instances in a process~\cite{MARTIN2020101463}. Outside these time intervals, the resource is considered to be off-duty, and thus not available to start an activity instance.

Since resources in a business process tend to operate according to daily or weekly cycles (herein called \emph{circadian cycles}), it is convenient to model an availability calendar by referring to time intervals within a day (e.g. 10:00-11:00) and to days within a week (e.g. Mondays). This leads us to the notion of\emph{circadian time-slot} (or \emph{time-slot} for short). For example, the time-slot ``Mondays 10:00-11:00'' describes a set of time intervals. Each of these time-intervals consists of a set of time points such that the day of the week is Monday, and the time of the day is between 10:00 and 11:00.

In line with previous work on resource availability calendars~\cite{LopezPintadoD23}, we use quarter-hour (15-minute periods) as the time granularity for availability calendars. In other words, we decompose a day into quarter-hours, e.g. 00:00-00:15, 00:15-00:30 and so on until 23:45-24:00 (95 granules per day). This is a sufficiently small granularity to capture typical work shifts, but not too small to potentially overfit. Indeed, if we modelled availability at the granularity of the minute or lower, we could end up with calendars containing time-slots such as Mondays 00:00:00-00:00:20, which would mean that a resource is on-duty for 20 seconds at the start of the day, and then goes off-duty -- a situation that is not typical for resources in a business process.

With this convention, we formally define a resource availability calendar as follows.

\begin{definition}[\textbf{Resource Availability Calendar}]\label{eq1} 
A \emph{circadian time instant} \(t\) is a tuple \((day, sec)\), where \(day \in [0..6]\) denotes the day of the week (Sunday = 0, Saturday = 6), and \(sec \in [0, 86400]\) represents the number of seconds elapsed since the start of that day. 

A \emph{time slot}($\tau$) is a tuple \((day, start, end)\), where \(day \in [0..6]\) specifies the day of the week, and \(start, end \in [0..95]\)\footnote{Note that there are 96 quarter-hours on a day, herein denoted with numbers 0 to 95.} denote the start and end times in units of 15-minute intervals since the beginning of the day. Each interval corresponds to \(900\) seconds denoting the number of seconds in a quarter of an hour. The start and end times are such that \(start \le end\). 

A \emph{calendar} is a set of non-overlapping time slots $\{(d_1, s_1, e_1), (d_2, s_2, e_2), \dots \}$.

Given a resource $R$, the \emph{availability calendar} \(A(R)\) of $R$ is a calendar consisting of the time-slots when resource $R$ is available to perform work (i.e.\ to start an activity instance).

The following boolean function allows us to determine if a resource is available (on duty) at a given circadian time point, given the resource's availability calendar.

\[
A(R, t) =
\begin{cases} 
true, & \text{if there exists a timeslot($\tau$) } (day_{\mathrm{ts}}, start_{\mathrm{ts}}, end_{\mathrm{ts}}) \in A(R) \\  
   & \text{such that } day_t = day_{\mathrm{ts}} \text{ and } 900 \cdot start_{\mathrm{ts}} \le sec_t \le 900 \cdot end_{\mathrm{ts}}, \\
false, & \text{otherwise,}
\end{cases}
\tag{1}
\]

\end{definition}

The proposed approach for predicting case suffixes includes resource availability features. These features are extracted from the availability calendars of the resources that are referenced in the event log. Specifically, for each resource, we discover a resource availability calendar using the method reported in~\cite{LopezPintadoD23}.  
Using the discovered resource availability calendar \(A(R)\) for a resource \(R\) and given a cutoff point \( t_{\text{current}} \), we extract a feature vector, herein referred to as ${R}_{\text{avail}}$, consisting of four features: Time Until Next On-Duty, Time Until Next Off-Duty, Time Since Last On-Duty, and Time Since Last Off-Duty. 
These features are formally defined below.

\begin{definition}[Time Until Next On-Duty]
Given a cut-off point \( t_{\text{current}} \) and a resource \( R \), 
the \textit{Time Until Next On-Duty} is the time remaining from \( t_{\text{current}} \) until resource \( R \) next becomes on duty, taking into consideration the availability calendar of this resource. If the resource is already on duty as of time \( t_{\text{current}} \), the \textit{Time Until Next On-Duty} is 0. Otherwise, it is the time difference between the current time and the next scheduled on-duty time for the resource. 

To compute \textit{Time Until Next On-Duty}, we use two functions DayOfWeek and TimeOfDay, which given an absolute time point, return (respectively) the day of the week and the time since the start of the day, measured in seconds. For example DayOfWeek(`2026-01-05 00:00:25') = 6 (since `2026-01-10' is a Saturday) and TimeOfDay(`2026-01-05 00:00:25') = 25, since there are 25 elapsed seconds from 00:00:00 to 00:00:25.

Given a time-slot $\tau = (day, start, end)$ of a calendar, let DayOfWeek($\tau$) = $\mathit{day}$, i.e. DayOfWeek($\tau$) be the day of the week of time-slot $\tau$. 
Further, let TimeOfStartDay($\tau$) = $15 \times 60 \times \textit(start)$ be the number of seconds since midnight and the start of time-slot $\tau$. Note that $\textit{start}$ is the number of quarter-hours since the start of the day and the start of the time-slot. Note also that the factor $15 \times 60$ captures the number of seconds in one quarter-hour.

Now, given a time point \( t_{\text{current}} \), we define a function $W$ that calculates the time difference (in seconds), between \( t_{\text{current}} \) and the start of the week when \( t_{\text{current}} \) occurs.

\[ W(t_{\text{current}}) = 24 \times 60 \times 60 \times DayOfWeek(t_{\text{current}}) + TimeOfDay(t_{\text{current}})
\]

Note that $24 \times 60 \times 60 = 86400$ is the number of seconds in a day.

Similarly, given a time slot  $\tau$, we define $W_{s}(\tau)$ as the time elapsed between the start of the week and the start of $\tau$.

\[ W_{s}(\tau) = 24 \times 60 \times 60 \times DayOfWeek(\tau) + TimeOfStartDay(\tau)
\]

Given a time-slot having start timestamp $\tau$, we can now define the time elapsed from  $t_{\text{current}}$  to $\tau_s$ as follows:

\[ TimeElapsed(t_{\text{current}}, \tau) = ( W_{s}(\tau) - W(t_{\text{current}}) + 604800) mod 604800  
\]

Note that $604800$ is the number of seconds in a week. We add this term to the difference between $t_{\text{current}}$ and $\tau$ and then compute the modulo of the result to account for the fact that circadian time is circular, e.g.\ the elapsed time from Saturday at 23:59:59 and Sunday at 00:00:01 is not $1 - 604799 = -604798$ but rather $(1 - 604799 + 604800) \; \mathit{mod} \;  604800 = 2 \; \mathit{mod} \; 604800 = 2$.

Given an availability calendar A(R), consisting of a set of timeslots, we define the time until the next time slot as the minimum $\mathit{TimeElapsed}$ from $t_{\text{current}}$ to any time-slot $\tau \in A(R)$. However, we need to take into account that if the resource is on-duty as of time $t_{\text{current}}$, then the time until next on-duty is zero. These observations lead us to the following formula:

\begin{equation}
\begin{aligned}
& T_{\text{next\_on\_duty}}(R, t_{\text{current}}) = 
\begin{cases}
0
& \text{if } A(R, t_{\text{current}}) \text{ is true} \\
\min \bigl( \{ \mathrm{TimeElapsed}(t_{\text{current}}, \tau)\\
\mid \tau \in A(R) \} \bigr),
& \text{otherwise}
\end{cases}
\end{aligned}
\tag{2}
\end{equation}

\end{definition}

For instance, if a resource is scheduled to start activity instance at 3:00 PM (i.e., \( \tau^{\text{next}} = 15:00:00 \)) and the current time is 2:30 PM (i.e., \( t_{\text{current}} = 14:30:00 \)), the \textit{Time Until Next On-Duty} will be 30 minutes, or 1800 seconds. On the other hand, if the current time is 4:30 PM and the resource is already on duty, the \textit{Time Until Next On-Duty} will return 0.




\begin{definition}[Time Until Next Off-Duty]
Given a cut-off point \( t_{\text{current}} \) and a resource \( R \), 
the \textit{Time Until Next Off-Duty} is the time remaining from 
\( t_{\text{current}} \) until resource \( R \) next becomes off duty, 
taking into consideration the availability calendar of this resource. 
If the resource is already off duty as of time \( t_{\text{current}} \), 
the \textit{Time Until Next Off-Duty} is 0. Otherwise, it is the time 
difference between the current time and the scheduled end of the 
availability interval in which the resource is currently on duty.

As in Definition~2, we rely on the functions DayOfWeek and TimeOfDay 
to decompose absolute time points, and we use the function \( W \) to 
map time points to seconds elapsed since the start of the week.

Given a time-slot \( \tau = (day, start, end) \) of an availability 
calendar, let DayOfWeek(\(\tau\)) \(= day\), and let
\[
\text{TimeOfEndDay}(\tau) = 15 \times 60 \times \textit{end},
\]
which represents the number of seconds since midnight corresponding 
to the end of the time-slot. We then define
\[
W_{e}(\tau) = 24 \times 60 \times 60 \times \text{DayOfWeek}(\tau) 
+ \text{TimeOfEndDay}(\tau),
\]
i.e., the number of seconds elapsed between the start of the week and 
the end of time-slot \( \tau \).

The time elapsed from \( t_{\text{current}} \) to the end of a time-slot 
\( \tau \) is defined analogously as
\[
\text{TimeElapsed}(t_{\text{current}}, \tau) =
\bigl( W_{e}(\tau) - W(t_{\text{current}}) + 604800 \bigr) \bmod 604800,
\]
where \( 604800 \) is the number of seconds in a week.

Let \( A(R) \) denote the availability calendar of resource \( R \), and 
let \( A(R, t_{\text{current}}) \) indicate whether the resource is on 
duty at time \( t_{\text{current}} \). The \textit{Time Until Next 
Off-Duty} is then defined as:

\begin{equation}
T_{\text{next\_off\_duty}}(R, t_{\text{current}}) = \\
\begin{cases}
0,
& \text{if } A(R, t_{\text{current}}) \text{ is false}, \\
\displaystyle
\min \{ \mathrm{TimeElapsed}(t_{\text{current}}, \tau)\\
\mid \tau \in A(R)\},
& \text{otherwise.}
\end{cases}
\tag{3}
\end{equation}

For instance, if a resource is scheduled to remain on duty until 
6:00~PM (i.e., the end of the current availability interval is 
\( \tau = 18{:}00{:}00 \)) and the current time is 4:30~PM 
(i.e., \( t_{\text{current}} = 16{:}30{:}00 \)), the 
\textit{Time Until Next Off-Duty} is 1.5 hours, or 5400 seconds. 
On the other hand, if the current time is 7:00~PM and the resource 
is already off duty, the \textit{Time Until Next Off-Duty} returns 0.
\end{definition}

\begin{definition}[Time Since Last On-Duty]
Given a cut-off point \( t_{\text{current}} \) and a resource \( R \), 
the \textit{Time Since Last On-Duty} measures the time that has elapsed 
since resource \( R \) last became on duty, taking into consideration 
the availability calendar of the resource. If the resource is currently 
off duty as of time \( t_{\text{current}} \), the \textit{Time Since Last 
On-Duty} is 0. Otherwise, it is the time difference between the current 
time and the start of the availability interval in which the resource 
is currently on duty.

As in the previous definitions, we rely on the functions DayOfWeek and 
TimeOfDay to decompose absolute time points, and we use the function 
\( W \) to map time points to seconds elapsed since the start of the 
week.

Given a time-slot \( \tau = (day, start, end) \) of an availability 
calendar, let DayOfWeek(\(\tau\)) \(= day\), and let
\[
\text{TimeOfStartDay}(\tau) = 15 \times 60 \times \textit{start},
\]
which represents the number of seconds since midnight corresponding 
to the start of the time-slot. We then define
\[
W_{s}(\tau) = 24 \times 60 \times 60 \times \text{DayOfWeek}(\tau)
+ \text{TimeOfStartDay}(\tau_s),
\]
i.e., the number of seconds elapsed between the start of the week and 
the start of time-slot \( \tau \).

The time elapsed from the start of a time-slot \( \tau \) to 
\( t_{\text{current}} \) is defined as
\[
\text{TimeElapsed}(\tau, t_{\text{current}}) =
\bigl( W(t_{\text{current}}) - W_{s}(\tau) + 604800 \bigr) \bmod 604800,
\]
where \( 604800 \) is the number of seconds in a week.

Let \( A(R) \) denote the availability calendar of resource \( R \), and 
let \( A(R, t_{\text{current}}) \) indicate whether the resource is on 
duty at time \( t_{\text{current}} \). The \textit{Time Since Last 
On-Duty} is then defined as:



\begin{equation}
T_{\text{since\_last\_on\_duty}}(R, t_{\text{current}}) = \\
\begin{cases}
0,
& \text{if } A(R, t_{\text{current}}) \text{ is false}, \\
\displaystyle
\min \{ \mathrm{TimeElapsed}(\tau, t_{\text{current}})\\
\mid \tau \in A(R)\},
& \text{otherwise.}
\end{cases}
\tag{4}
\end{equation}

For instance, if a resource started its current availability interval 
at 1{:}00~PM (i.e., \( \tau = 13{:}00{:}00 \)) and the current time is 
4{:}00~PM (i.e., \( t_{\text{current}} = 16{:}00{:}00 \)), the 
\textit{Time Since Last On-Duty} is 3 hours, or 10{,}800 seconds. If the 
current time falls outside all availability intervals, the value 
returned is 0.
\end{definition}





\begin{definition}[Time Since Last Off-Duty]
Given a cut-off point \( t_{\text{current}} \) and a resource \( R \), 
the \textit{Time Since Last Off-Duty} measures the time that has elapsed 
since resource \( R \) last became off duty, taking into consideration 
the availability calendar of the resource. If the resource is currently 
on duty as of time \( t_{\text{current}} \), the \textit{Time Since Last 
Off-Duty} is 0. Otherwise, it is the time difference between the current 
time and the end of the availability interval immediately preceding the 
current off-duty period.

As in the previous definitions, we rely on the functions DayOfWeek and 
TimeOfDay to decompose absolute time points, and we use the function 
\( W \) to map time points to seconds elapsed since the start of the 
week.

Given a time-slot \( \tau = (day, start, end) \) of an availability 
calendar, let DayOfWeek(\(\tau\)) \(= day\), and let
\[
\text{TimeOfEndDay}(\tau) = 15 \times 60 \times \textit{end},
\]
which represents the number of seconds since midnight corresponding 
to the end of the time-slot. We then define
\[
W_{e}(\tau) = 24 \times 60 \times 60 \times \text{DayOfWeek}(\tau)
+ \text{TimeOfEndDay}(\tau),
\]
i.e., the number of seconds elapsed between the start of the week and 
the end of time-slot \( \tau \).

The time elapsed from the end of a time-slot \( \tau \) to 
\( t_{\text{current}} \) is defined as
\[
\text{TimeElapsed}(\tau, t_{\text{current}}) =
\bigl( W(t_{\text{current}}) - W_{e}(\tau) + 604800 \bigr) \bmod 604800,
\]
where \( 604800 \) is the number of seconds in a week.

Let \( A(R) \) denote the availability calendar of resource \( R \), and 
let \( A(R, t_{\text{current}}) \) indicate whether the resource is on 
duty at time \( t_{\text{current}} \). The \textit{Time Since Last 
Off-Duty} is then defined as:


\begin{equation}
T_{\text{since\_last\_off\_duty}}(R, t_{\text{current}}) = \\
\begin{cases}
0,
& \text{if } A(R, t_{\text{current}}) \text{ is true}, \\
\displaystyle
\min \{ \mathrm{TimeElapsed}(\tau, t_{\text{current}})\\
\mid \tau \in A(R)\},
& \text{otherwise.}
\end{cases}
\tag{5}
\end{equation}

For instance, if a resource ended its most recent availability interval 
at 5{:}00~PM (i.e., \( \tau = 17{:}00{:}00 \)) and the current time is 
7{:}00~PM (i.e., \( t_{\text{current}} = 19{:}00{:}00 \)), the 
\textit{Time Since Last Off-Duty} is 2 hours, or 7{,}200 seconds. If the 
resource is currently on duty, the value returned is 0.
\end{definition}

Algorithm~\ref{alg:enhance_traceprefix} computes resource availability features by querying precomputed resource availability calendars $C_r$ obtained following~\cite{LopezPintadoD23}. For each trace prefix $\traceprefix \subseteq \sigma_c$, availability is evaluated at the current decision time $T^{\text{current}}$, defined as the start time of the active activity instance $e_{c,k}$.The algorithm first identifies the set of resources $\mathcal{R}^{a_k} = \mathcal{R}(a_{c,k})$ that can execute the activity $a_{c,k}$ (line~\texttt{30}). For each eligible resource $r \in \mathcal{R}^{a_k}$, four temporal availability features are extracted from the corresponding calendar $C_r$ at time $T^{\text{current}}$: time to next on-duty, time since last on-duty, time to next off-duty, and time since last off-duty (lines~\texttt{31--35}). These features are aggregated across all eligible resources by computing the minimum value per feature dimension (lines~\texttt{36--38}), yielding an availability vector $\bar{R}_{\text{avail}}$. Finally, this vector is appended to the active activity instance $e_{c,k}$, enriching it with resource availability information used by case suffix prediction models.


\subsubsection{Feature Vector Construction} 
Our approach uses three feature types: intra-case features from individual cases, resource contention features from all active cases and resource availability features extracted from resource calendars. These are encoded into feature vectors for each \traceprefix, combining all contexts to form input sequences.

\paragraph{Feature Encoding:} 
To represent the input data for the LSTM, we encode intra-case, resource contention (${R}_{\text{cont}}$) and resource availability (${R}_{\text{avail}}$) features in a structured and consistent manner. Intra-case features include attributes specific to each event in a case, such as the activity label, inter-start time, and processing time. Activity labels, being categorical, are embedded using trainable embedding layers rather than one-hot encoding. These embeddings are learned jointly with the LSTM network during supervised training, allowing the model to capture task-specific semantic relationships between activities and efficiently handle a larger set of categories, following the approach in~\cite{0001DR19}. The temporal features, inter-start time and processing time are treated as continuous variables and normalized to the [0,1] range using min-max normalization based on the training data, as illustrated in Table~\ref{ngram}.

\begin{table}[htbp]
\centering
\footnotesize
\renewcommand{\arraystretch}{0.8} 

\setlength{\tabcolsep}{8pt}       
\caption{N-grams snippet for Table.~\ref{instace_log}.}

\begin{tabular}{@{}c c c c@{}}
\toprule
\textbf{Case} & \textbf{Act\(_{\text{Seq}}\)} & \(\boldsymbol{\delta^{\text{start}}}\) & \(\boldsymbol{\delta^{\text{proc}}}\) \\
\midrule
\multirow{4}{*}{1} & [0000] & 0  & 5  \\
                   & [0001] & 10 & 5  \\
                   & [0012] & 10 & 40 \\
                   & [0123] & 45 & 5  \\
\midrule
\multirow{2}{*}{2} & [0000] & 0  & 5         \\
                   & [0001] & 10 & \(\emptyset\) \\
\midrule
\multirow{3}{*}{3} & [0000] & 0  & 5  \\
                   & [0001] & 10 & 5  \\
                   & [0012] & 10 & 40 \\
  
\bottomrule\\

\end{tabular}

\label{ngram}
\end{table}

Resource contention features (i.e.  WIP, resource utilization \& demand rate ($\lambda$)) capture the broader system context, while resource availability features (i.e. ${R}_{\text{avail}}$) represent the current availability of resources. These features are updated at each time step and normalized using min-max scaling to ensure standardization of data. Each feature is aligned with its corresponding event to maintain temporal coherence and provide accurate contextual information at every step.

\paragraph{Sequence Construction:}

To construct input sequences and target labels for training, we adopted a fixed-size n-gram approach for each \traceprefix{} in \activityinstancelog, as outlined in previous studies~\cite{0001DR19,TaxVRD17}. Since variable-length prefixes cannot be directly used as inputs, the n-gram method enables standardization of the temporal dimensionality of the input data hence, capturing sub-sequence patterns of activity instances in \traceprefix{}. Each n-gram sequence incorporates intra-case features (e.g., activity, inter-start time, processing time), ${R}_{\text{cont}}$ features (e.g., WIP, resource utilization, \(\lambda\)) and ${R}_{\text{avail}}$ features ensuring the model captures both case-specific and system-wide process dynamics. Sequences are generated independently for each \(\traceprefix{}\), maintaining consistency across all instances, as shown in Table.~\ref{ngram}.

\subsubsection{Model Architecture Design and Training} \label{model_training}
Existing SM architectures predicts all aspects of the next activity instance in a trace prefix \traceprefix{}, i.e., the next activity, its inter-start time, and processing time jointly.  However, this joint prediction approach lacks flexibility in scenarios where only partial predictions are required. For example, if the activity and its start time are already known, as is the case for ongoing activities, it is often enough to predict only the processing time. In SM-based architectures, this is difficult because all aspects of the next activity instance are predicted together, and the model cannot easily make just one specific prediction. To address this, we propose a BiLSTM-based Multi-model Predictive Learning Architecture (MM)~\cite{GunnarssonBW23}, which leverages bidirectional LSTMs to predict the next activity, inter-start time, and processing time for a trace prefix \traceprefix{}. BiLSTM analyzes event sequences bidirectionally, capturing activity dependencies, temporal variations, and overlaps.


MM combines categorical and continuous inputs using pre-trained embeddings, concatenated features, and BiLSTM layers to model sequential dependencies. It follows a modular design with three independent BiLSTM predictors: \(\alpha\) for next activity, \(\beta\) for inter-start time (\(\interstarttime\)), and \(\gamma\) for processing time (\(\interprocessingtime\)), as shown in Fig.~\ref{offlinefig}. Each model is trained separately on \(\activityinstancelog_{train}\) to capture distinct aspects of the process.

To train the next activity predictor \(\alpha\), we extract $\traceprefix{}$'s from \(\activityinstancelog_{train}\) and construct intra-case, ${R}_{\text{cont}}$ and ${R}_{\text{avail}}$ feature vectors using Alg.~\ref{alg:enhance_traceprefix} to capture both process dynamics and resource context. These features are encoded into input sequences with target labels to train \(\alpha\). For the inter-start time predictor \(\beta\), we use a similar feature set, but the model additionally takes the actual next activity \(a_{c,k+1}\) from \(\traceprefix_{k+1}\) as input to learn accurate temporal dependencies. Fixed-length input sequences and corresponding labels are generated for training. The processing time predictor \(\gamma\) also uses similar features, but focuses on estimating the duration of the next activity \(a_{c,k+1}\), incorporating both \(a_{c,k+1}\) and its inter-start time \(\interstarttimeplusone\). This enables \(\gamma\) to capture temporal patterns and case-specific variations for accurate processing time prediction.

\subsubsection{Output}

The offline phase produces three independently trained BiLSTM predictors. Given a trace prefix \traceprefix{}, the activity predictor \(\alpha\) outputs a probability distribution over possible next activities. Conditioned on \traceprefix{} and the next activity \(a_{c,k+1}\), the inter-start time predictor \(\beta\) estimates \(\interstarttimeplusone\), while the processing time predictor \(\gamma\) estimates \(\interprocessingtimeplusone\) based on \traceprefix{}, \(a_{c,k+1}\), and \(\interstarttimeplusone\). These predictors are selectively used in the online phase depending on the available runtime information.

\subsection{Online Phase}
\begin{figure}[h]
    \centering
    \includegraphics[width=\linewidth]{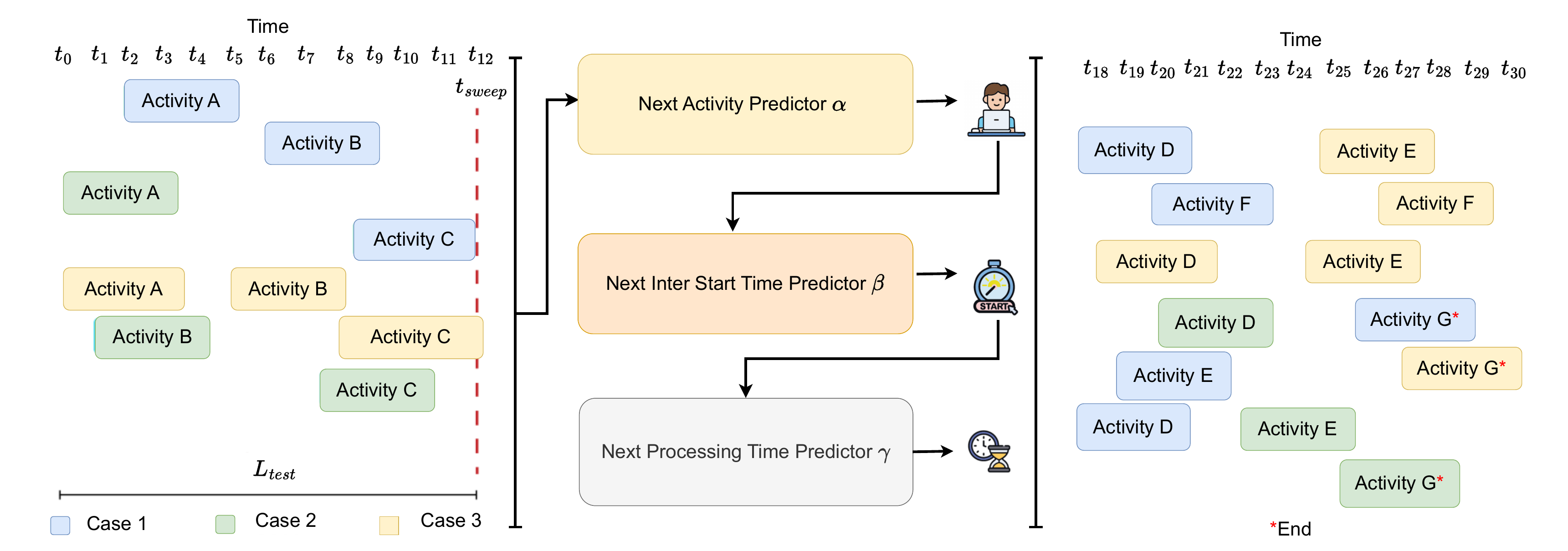}
    \caption{Online Phase}
    \label{onlinefig}
\end{figure}

The online phase (Fig.~\ref{onlinefig}) uses the predictive models (\(\alpha, \beta, \gamma\)) trained during the offline phase to predict case suffixes using a sweep-line-based method, wherein the suffixes of all ongoing cases in the process are predicted in lockstep rather than in isolation.  Alg.~\ref{alg:sweep_line_with_predictions} details the online phase, iteratively processing ongoing cases in chronological order of their predicted start timestamps.
\subsubsection{Input} \label{online_input}

In the online phase, Alg.~\ref{alg:sweep_line_with_predictions} takes as input a log \activityinstancelog{}, a cutoff time point \( t_{\text{cutoff}} \), and three models \(\alpha, \beta, \gamma\). The cutoff time \( t_{\text{cutoff}} \) marks the beginning of the online phase, from which future activity instances are predicted in lockstep.

To reflect the system's state at \( t_{\text{cutoff}} \), the log \activityinstancelog{} is modified by removing all activity instances that start after the \( t_{\text{cutoff}} \). For activity instances that begin before but end after \( t_{\text{cutoff}} \), the end timestamp is set to \(\emptyset\) to indicate the activity was still ongoing. The resulting set of activity instance traces \(\activityinstancetraces{}\) forms a collection of incomplete trace prefixes, denoted \(\traceprefix{}_{\emptyset}\), each representing a partial sequence of completed and ongoing activities.

Each \(\traceprefix{}_{\emptyset}\) is linked to a case identifier and contains the activity name, start timestamp, and if available end timestamp for each instance in \(\activityinstancetraces{}\), as illustrated in Table~\ref{instace_log}. These prefixes collectively form the \emph{test log} (\(L'\)) (line~\texttt{1}). Providing \(\traceprefix{}_{\emptyset}\) as input to MM offers the historical context required to predict the next activity and its expected start and end timestamps for each ongoing case.

To predict suffixes of all ongoing cases in  lock step as of \( t_{\text{cutoff}} \), the Alg.~\ref{alg:sweep_line_with_predictions} begins by initializing the \(\TemporalSweep{}\) at \(t_{\text{cutoff}}\) (line~\texttt{2}), which serves as the starting point for the sweep-line algorithm to begin completing the incomplete trace prefixes (\(\traceprefix{}_{\emptyset}\)). MM integrates three models \(\alpha, \beta,  \gamma\) in to a sweep-line based simulation algorithm to sequentially predict the next activity, its inter-start time and its processing time for a given incomplete trace prefix $\traceprefix{}_{\emptyset}$. 

While \TemporalSweep{} is less than the maximum start timestamp in \(\activityinstancelog'\), the algorithm extracts incomplete trace prefixes (\(\traceprefix_{\emptyset}\)) as of \TemporalSweep{} (line \texttt{4}). Each \(\traceprefix_{\emptyset}\) includes traces where the last activity has started but not yet finished. Lines (\texttt{5-10}) handle such cases by first completing these ongoing activities. For each prefix \( \traceprefix_{c,k-1} \), intra-case, ${R}_{\text{cont}}$ \& ${R}_{\text{avail}}$ features  are extracted using Alg.~\ref{alg:enhance_traceprefix} (line \texttt{7}) and passed to the prediction model \( \gamma \) to estimate the remaining processing time (line \texttt{8}). The end timestamp is then computed by adding the predicted \(\interprocessingtime{}\) to the start time of \( \text{a}_{c,k} \) (line \texttt{9}). To support concurrent activities, the algorithm applies this prediction to all activity instances in the prefix \( P_c \) with missing end timestamps, rather than only the most recent one, ensuring consistent suffix prediction in the presence of parallelism.

With complete activity instances, \(\traceprefix_{c}\) is enhanced with intra-case, ${R}_{\text{cont}}$ \& ${R}_{\text{avail}}$ features using Alg.~\ref{alg:enhance_traceprefix} (line \texttt{12}). For each enhanced \(\traceprefix_{c}\) (lines \texttt{13-15}), MM sequentially applies its models, where \( \alpha \) predicts the next activity instance \( \text{act}_{c,k+1} \) given trace \( \traceprefix_{c} \). \( \beta \) predicts its inter-start time \( \interstarttimeplusone \) given \( \text{a}_{c,k+1} \) and \( \traceprefix_{c} \) and \( \gamma \) predicts its processing time given \( \text{a}_{c,k+1} \), \( \interstarttimeplusone \), and \(\traceprefix_{c}\).

The start timestamp is obtained by adding \( \interstarttimeplusone \) to the previous activity’s start timestamp, while the end timestamp is calculated by adding the processing time to the start timestamp (line \texttt{16-17}). The predicted activity instances are then appended to \(\traceprefix_{c}\), extending the ongoing cases (line \texttt{18}). The updated \(\traceprefix_{c}\) is subsequently added to \(\activityinstancelog'\) for the next prediction cycle. The algorithm then advances  to the next event time by selecting the minimum start timestamp in \(\activityinstancelog'\) that is greater than the current  and not associated with an end-of-trace (EOT) activity (line~\texttt{20}). If no such timestamp exists, the loop terminates. This process repeats until all cases are completed, meaning the next predicted activity for every trace is EOT.

\subsubsection{Output}
The output of the online phase is a completed activity instance log \(L'\), where each incomplete trace prefix observed at \(t_{\text{cutoff}}\) is extended with a predicted case suffix using the models \(\alpha\), \(\beta\), and \(\gamma\) trained during the offline phase. The predicted suffixes consist of a sequence of future activity instances, each associated with a predicted activity label and corresponding temporal information. Resource prediction for activity instances in the predicted suffixes is not performed, as prior work \cite{0001DR19} indicates limited accuracy at this level of granularity; instead, higher-level abstractions such as roles have been shown to be more informative, although role prediction is beyond the scope of this work. Overall, the resulting log \(L'\) represents a complete and temporally consistent extension of the test log and is directly comparable to a fully observed activity instance log.

\begin{algorithm}[htbp]
\caption{MM based Sweep Line Algorithm}
\label{alg:sweep_line_with_predictions}
\KwIn{Models $\alpha$, $\beta$, $\gamma$; activity instance log $L$; cutoff $t_{\text{cutoff}}$}
\KwOut{Completed log $L'$ with predicted suffixes}

Let $L' \leftarrow \texttt{Extract}(L, t_{\text{cutoff}})$\;

Let $t_{\text{sweep}} \leftarrow t_{\text{cutoff}}$\;


\While{$t_{\text{sweep}} \leq \max(T^{\text{start}} \text{ in } L')$}{
    Let $P_{\emptyset} \leftarrow$  $\texttt{Prefix}(L', t_{\text{sweep}})$\;
    

    \ForEach{$P_c \in P_{\emptyset}$}{
        \ForEach{$e = (c, a, T^{\text{start}}, T^{\text{end}}) \in P_c$ \textbf{where} $T^{\text{end}} = \emptyset$}{
            Enhance $P_c$ with intra, ${R}_{\text{cont}}$ \& ${R}_{\text{avail}}$ features using Alg.~\ref{alg:enhance_traceprefix}\;
            Let $\delta^{\text{proc}} \leftarrow \gamma(P_c)$\;
            $T^{\text{end}} \leftarrow T^{\text{start}} + \delta^{\text{proc}}$\;
            $e \leftarrow (c, a, \delta^{\text{proc}}, T^{\text{start}}, T^{\text{end}})$\;
        }
    }

    \ForEach{$P_c \in P_{\emptyset}$}{
        Enhance $P_c$ with intra, ${R}_{\text{cont}}$ \& ${R}_{\text{avail}}$ features using Alg.~\ref{alg:enhance_traceprefix}\;

        Let $a_{k+1} \leftarrow \alpha(P_c)$\;
        Let $\delta^{\text{start}}_{k+1} \leftarrow \beta(P_c, a_{k+1})$\;
        Let $\delta^{\text{proc}}_{k+1} \leftarrow \gamma(P_c, a_{k+1}, \delta^{\text{start}}_{k+1})$\;

        $T^{\text{start}}_{k+1} \leftarrow T^{\text{start}}_k + \delta^{\text{start}}_{k+1}$\;
        $T^{\text{end}}_{k+1} \leftarrow T^{\text{start}}_{k+1} + \delta^{\text{proc}}_{k+1}$\;

        Append $(c, a_{k+1}, \delta^{\text{start}}_{k+1}, \delta^{\text{proc}}_{k+1}, T^{\text{start}}_{k+1}, T^{\text{end}}_{k+1})$ to $P_c$\;
    }

    Update $L'$ with modified prefixes $P_c$\;
    

    
    $t \leftarrow \left\{ T^{\text{start}} \in L' \;\middle|\; T^{\text{start}} > t_{\text{sweep}} \text{ and activity} \neq \text{EOT} \right\}$\;

    \textbf{if} $t \neq \emptyset$ \textbf{then} \\
    \hspace*{1em} $t_{\text{sweep}} \leftarrow \min(t)$ \\
    \textbf{else} \\
    \hspace*{1em} \textbf{break}
}
\Return $L'$
\end{algorithm}

\section{Evaluation}\label{sec4}
Previous methods use \emph{Single Model Architectures} (SM), where a single model is trained to predict both the next activity and its timestamp. In contrast, as discussed in Sec.~\ref{sec3}, our MM approach applies separate models sequentially to predict the next activity, its inter-start time, and its processing time. This modular design offers greater flexibility, especially when the activity and start time are already known at a given \TemporalSweep{}, which enables more accurate processing time predictions. However, it remains to be seen whether this modularity affects predictive performance. The evaluation below addresses the following questions:

\begin{enumerate}

\item \textit{\textbf{EQ1}: How does MM compare to SM in predicting the suffix of ongoing cases in a prefix log?}

\begin{enumerate}
    
    \item \textit{\textbf{EQ1(a)} In terms of control flow prediction?}
    We hypothesize that \emph{MM’s flexibility} -- enabling the integration of the most suitable control flow prediction model for each event log through targeted training -- enables it to outperform SM-based baselines. We also expect that the \emph{resource contention features} (resource utilization, demand, and capacity) will \emph{not} enhance control-flow accuracy, as they capture workload properties rather than execution order.

    \item \textit{\textbf{EQ1(b)} In terms of inter-start time prediction?}
    Since \emph{inter-start times} (i.e., time from the start of the current activity to the start of the next activity) are influenced by \emph{resource utilization, demand, and capacity}, we expect that incorporating \emph{resource contention features} into our \emph{sweep-line approach} will improve their prediction. Resource contention features provide needed information on workload and availability, which are determinants of waiting time.  
    Furthermore, we hypothesize that \emph{better control flow prediction} will further enhance inter-start time estimation. Since activities have varying waiting times, a better next-activity predictor improves inter-start time prediction and  hence start timestamp estimation.
    
    \item \textit{\textbf{EQ1(c)} In terms of processing time prediction?} Unlike inter-start times, \emph{processing times} are \emph{less dependent} on workload factors such as resource utilization, demand and capacity. Instead, they are primarily determined by \emph{the nature of the activity} itself. Consequently, we do \emph{not} expect resource contention features to improve processing time prediction. However, we anticipate that a \emph{more accurate next-activity prediction} could improve processing time estimates. Since activities have distinct processing times, accurate next-activity prediction enhances processing time estimation.
    
\end{enumerate}
\item \textit{\textbf{EQ2}: How does the inclusion of resource availability features influence MM’s predictive performance?} 
While EQ1 compares the predictive power of MM and SM, EQ2 compares MM and MM$_{avail}$ (where MM$_{avail}$ represents the model with resource-availability features), isolating the impact of \emph{resource-availability features}. Specifically, we assess whether incorporating information on \emph{resource availability} enable improved predictions across control flow, inter-start time, and processing time tasks.  
We hypothesize that incorporating \emph{resource availability features} can improve predictive performance across all three tasks. When information about which resources are available at a given time is included, the model can better infer which activities are more likely to occur next, thereby enhancing \emph{control flow prediction}. Moreover, since the start of the next activity often depends on when a suitable resource becomes available, these features are expected to substantially improve \emph{inter-start time prediction}. Finally, as both the control-flow and inter-start times are predicted more accurately, this most likely yield more precise \emph{processing time predictions}.  
\end{enumerate}

\subsection{Experiment Setup}
To simulate real-life scenarios where models are trained on historical data and applied to ongoing cases, we represent each training and test instance as a pair consisting of a prefix and a suffix trace, denoted as $(\sigma \leq k, \sigma > k)$, where the prefix length $k$ is at least 1. To prevent data leakage, we implement a strict temporal split~\cite{VerenichDRMT19} when dividing the event log into training and testing sets. We define the duration of a log as the time elapsed between the earliest start timestamp (lower bound) and the latest recorded end timestamp (upper bound). The cutoff point $t_{\text{cutoff}}$ is determined as the timestamp where 80\% of the total process duration has elapsed. Cases that complete within the first 80\% of the timeframe are assigned to the training set, while the remaining 20\% are allocated to the test set. Additionally, cases that start before the cutoff but remain ongoing, along with their activity instances, are included in the test set. This ensures the test phase reflects real-world scenarios where predictions handle incomplete cases.

In SM-based experiments, activity instances that start and finish before the cutoff are excluded from the test set, as SM predicts the next activity, inter-start time, and processing time together but cannot handle ongoing instances. In contrast, MM includes ongoing instances since it can predict their processing time given its activity and start timestamp.

Experiments were conducted on a desktop with an NVIDIA RTX 3090 GPU, Intel Core i9 CPU, and 64 GB RAM. Training the three MM models ($\alpha$, $\beta$, $\gamma$) takes 2.5 hours for smaller logs (e.g., ACR, CFS) and 5-6 hours for larger ones (e.g., CVS, BPI2017W). During testing, the sweep-line approach predicts suffixes of ongoing cases at runtime. For BPI2017W, the approach generated 15,000 predicted events in 100 seconds at 150 events per second.

\subsubsection{Next Activity Prediction in the MM Architecture} A core component of the MM architecture is a model $\alpha$ that predicts the next activity in a case. Using an LSTM-based approach, the model generates multiple possible next activities, each assigned a probability score. A sampling method then selects the next activity. We evaluated MM’s accuracy with three sampling methods: Argmax, which selects the most probable activity~\cite{TaxVRD17}; Random Choice, which randomly selects an activity based on the probability distribution~\cite{0001DR19}; and Daemon Action, a heuristic-based method~\cite{AliDM24}. All three methods performed similarly, but Daemon Action produced the most consistent results across datasets. Therefore, we selected Daemon Action as the preferred sampling method for model $\alpha$.

\subsubsection{Hyperparameter Optimization}
We applied hyperparameter optimization within the sets of values in Table~\ref{tab:hparams} for selecting the most suitable three MM models ($\alpha$ for next activity, $\beta$ for inter-start time, $\gamma$ for processing time). Batch size is optimized for computational efficiency and gradient stability, while normalization methods like log normalization and max scaling improve convergence. The n-gram size (N\_size) ensures meaningful historical dependencies, and different LSTM layer sizes (L\_size) balance model complexity and training efficiency. Activation functions (Selu, Tanh) help maintain stable gradients, and optimizers (Nadam, Adam, SGD, Adagrad) are chosen for optimal convergence. The training set is split (80\% training, 20\% validation), and MM is trained with diverse hyperparameter combinations. We optimized hyperparameters using a random search over 50 iterations, which provides a practical balance between search space coverage and computational efficiency. Each model was trained for up to 200 epochs with early stopping(patience = 10) to prevent overfitting and reduce unnecessary training time. These settings are consistent with prior studies~\cite{0001DR19,AliDM24} and were empirically validated to ensure stable convergence. 

\begin{table}[htbp]
\centering
\caption{Hyperparameter settings for LSTM model}
\small
\begin{tabular}{lcc}
\toprule
\textbf{Parameter} & \textbf{Explanation} & \textbf{Search Space} \\
\midrule
Batch size & Number of samples per update & \{32, 64, 128\} \\
Normalization & Input scaling method & \{lognorm, max\} \\
Epochs & Training iterations & 200 \\
N\_size & Size of n-gram window & \{5, 10, 15, 20, 25\} \\
L\_size & LSTM layer size & \{50, 100, 150\} \\
Activation & Hidden layer activation & \{selu, tanh\} \\
Optimizer & Optimization algorithm & \{Nadam, Adam, SGD, Adagrad\} \\
\bottomrule
\end{tabular}
\label{tab:hparams}
\end{table}

During optimization, we evaluated performance using two metrics: Mean Absolute Error (MAE) \footnote{MAE is defined as \( \text{MAE} = \frac{1}{n} \sum_{i=1}^{n} |y_i - \hat{y}_i| \)} for \interstarttime{} and \interprocessingtime{} predictions; and Categorical Cross-Entropy Loss\footnote{ Categorical Cross-Entropy Loss is defined as \( \mathcal{L}_{\text{CE}} = - \sum_{i=1}^{n} \sum_{j=1}^{C} y_{ij} \log(\hat{y}_{ij}) \)} for next-activity prediction.

To evaluate BiLSTM configuration impact, we tested multiple MM variants with varying BiLSTM layer sizes and n-gram window sizes (see Table~\ref{tab:hparams}). Larger models marginally improved accuracy but increased training time and reduced throughput. The default configuration (100-unit layer size, 10 n-gram size, 64 batch size, 200 epochs, Adam optimizer) strikes a practical balance between accuracy and efficiency, offering an effective trade-off for predicting case suffixes in business processes.


\subsubsection{Measures of Goodness} 


To assess sequence similarity between predicted and ground-truth suffixes, we use the Damerau-Levenshtein (DL) distance. DL captures common edit operations -- insertions, deletions, substitutions, and transpositions -- that may occur between predicted and actual activity sequences~\cite{TaxVRD17}. We compute the DL distance for each predicted suffix against its corresponding ground-truth suffix and normalize it by the length of the longer sequence to ensure comparability across traces of varying lengths. A lower normalized DL score indicates higher similarity and better control-flow prediction accuracy.

Moreover, we use the Mean Absolute Error (MAE) to quantify the difference between predicted inter-start times and processing times (relative to the start time of the case) and the actual inter-start and processing times. A higher MAE between a set of predicted timestamps and a set of actual timestamps indicates a lower accuracy (thus lower MAE is better).

\subsubsection{Datasets} We evaluate our approach using nine event logs\footnote{All logs, including a mix of large- and small-scale processes, are available at supplementary material \href{https://figshare.com/s/cd6a11b081d80e3a9a28}{https://shorturl.at/PM3aV}}, selected for their diverse control-flow structures and temporal characteristics. Table~\ref{tab:event_log_stats} reports for each log the number of cases(\#Cases), activity instances (\#Act-Ins), and unique activities (\#Act). (\#Act/Case) is the average number of activities per case. CV Len and CV Dur  denotes the coefficient of variation in case length and duration respectively\footnote{The coefficient of variation (CV) is computed as: $\text{CV} = \frac{\sigma}{\mu}$, where $\sigma$ is the standard deviation and $\mu$ is the mean of case length or duration.}. Avg.Dur and Max.Dur shows average and maximum case durations in days. 

\begin{table}[htbp]
\centering
\footnotesize 
\setlength{\tabcolsep}{0.5pt} 
\caption{Event Log Statistics}
\label{tab:event_log_stats}
\begin{tabular}{l c c c c c c c c}

\hline
        \textbf{Log} &
\textbf{\#Cases} &
\textbf{\#Act-Ins} &
\textbf{\#Act} &
\textbf{\#Act/Case} &
\textbf{CV Len} &
\shortstack{\textbf{Avg. Dur}\\\textbf{(days)}} &
\shortstack{\textbf{Max. Dur}\\\textbf{(days)}} &
\textbf{CV Dur}   \\
        \midrule
        BPI17W            & 30,270           & 240,854                        & 8                        & 7.96               &       66.18\%        & 12.66                    & 286.07     & 706.70\%               \\
        BPI12W            & 8,616            & 59,302                         & 6                        & 6.88                &      104.11\%        & 8.91                     & 85.87        & 369.44\%             \\
       
        INS      & 1,182            & 23,141                         & 9                        & 19.58               &         74.27\%     & 70.93                    & 599.9            & 459.36\%         \\
        ACR  & 954          & 4,962                          & 16                       & 5.2                    &     32.13\%      & 14.89                    & 135.84                 & 157.25\%   \\
         MP      & 225              & 4,503                          & 24                       & 20.01                 &     93.78\%       & 20.63                    & 87.5                 & 130.40\%     \\

          CVS   & 10,000           & 103,906                        & 15                       & 10.39                 &     11.00\%       & 7.58                     & 21.0                & 642.64\%      \\
          
         CFS  & 1,000         & 21,221                         & 29                       & 26.53              &      54.60\%         & 0.83                     & 4.09                   & 83.52\%   \\

        CFM  & 2,000         & 44,373                         & 29                       & 26.57              &       55.41\%        & 0.76                     & 5.83                  & 83.82\%    \\

        P2P  & 608          & 9,119                          & 21                       & 15                   &      54.43\%       & 21.46                    & 108.31                 & 78.04\%   \\
     
        GOV  &   46300       &  259800                         &   142                     &              5.61      &     82.00\%       &  143.26                    & 1515.78                 & 124.00\%   \\
     
     WorkOrder  &   19800       &       149600                    &  24                      &       7.56             &      48.00\%       &  9.1                    & 112.23                 & 149.00\%   \\
     
     P2PFin  &   4000       &     30000                      &          16              &      7.58              &      67.00\%       & 58.09                    & 697.62                 & 151.00\%   \\
     \hline

\end{tabular}
\end{table}

All logs include start and end timestamps, as required by our approach. The \textbf{BPI12W} and \textbf{BPI17W} logs capture real-life financial processes from a Dutch institution, with BPI17W being a refined version of BPI12W.\footnote{\url{https://doi.org/10.4121/uuid:3926db30-f712-4394-aebc-75976070e91f}, \url{https://doi.org/10.4121/uuid:5f3067df-f10b-45da-b98b-86ae4c7a310b}} These logs represent human-performed activities with moderate process complexity. The \textbf{INS} log represents an insurance claims process and features long case durations and high variability. The \textbf{ACR} log, sourced from a Colombian university’s BPM system, models academic credential recognition, with medium trace lengths and a balanced activity structure. The \textbf{MP} (Manufacturing Production) log captures production operations from an ERP system, characterized by long traces and high activity diversity. The \textbf{GOV} event log records the execution of an approval application system of a government agency. The \textbf{Workorder} is an event log of a field services process at a utilities company. Field services is a type of process in which services are provided to a customer in response to a complaint registered by that customer.

We also include three synthetic logs that simulate real-life operational settings. The \textbf{CVS} log models a retail pharmacy process, based on the simulation in \textit{Fundamentals of Business Process Management}~\cite{DumasRMR18}, and serves as a large-scale training set. The \textbf{CFS} and \textbf{CFM} logs are anonymized datasets derived from a confidential process, representing small- and medium-scale versions of the same underlying workflow, both featuring high activity density and resource contention. Finally, the \textbf{P2P} (Purchase-to-Pay) log is a synthetic dataset of a procurement process, offering high structural complexity and longer case durations. Similarly, \textbf{P2PFin} is another anonymized payment process used at a financial institution. Table~\ref{tab:event_log_stats} summarizes the statistics of all logs used in the evaluation.

\subsection{Comparison of MM and SM Architectures}

\subsubsection{Baselines} For EQ1, the experiments compare the performance of our MM architecture against several variants of a Single Model architecture (SM) baselines adapted from Camargo et al.~\cite{0001DR19}. Specifically, we evaluate four SM variants: \textit{Shared Categorical} and \textit{Full Shared}, each tested with and without resource contention features. In the \textit{Shared Categorical} variant, only the embeddings of categorical inputs (e.g., activity labels) are shared across tasks, while the remaining layers are task-specific. In contrast, the \textit{Full Shared} variant employs a fully shared architecture, where both categorical and continuous inputs are processed jointly through the same network layers for all tasks. These configurations allow us to isolate the effects of architectural sharing and context-aware features on predictive performance.

To ensure a fair comparison, all models use the same feature encoding and sequence construction methodology described in (Sec.~\ref{offlinephase}). Intra-case features are computed individually per case, while resource contention features are derived by sequentially traversing all cases to capture dynamic workload conditions. Activities are encoded as categorical variables using low-dimensional embeddings, and input sequences are constructed via an n-gram strategy, ensuring standardized and consistent feature representation across models.

\subsubsection{Results and Interpretation} 
\begin{figure}[t]
    \centering
    \begin{subfigure}[t]{0.48\linewidth}
        \centering
        \includegraphics[width=\linewidth]{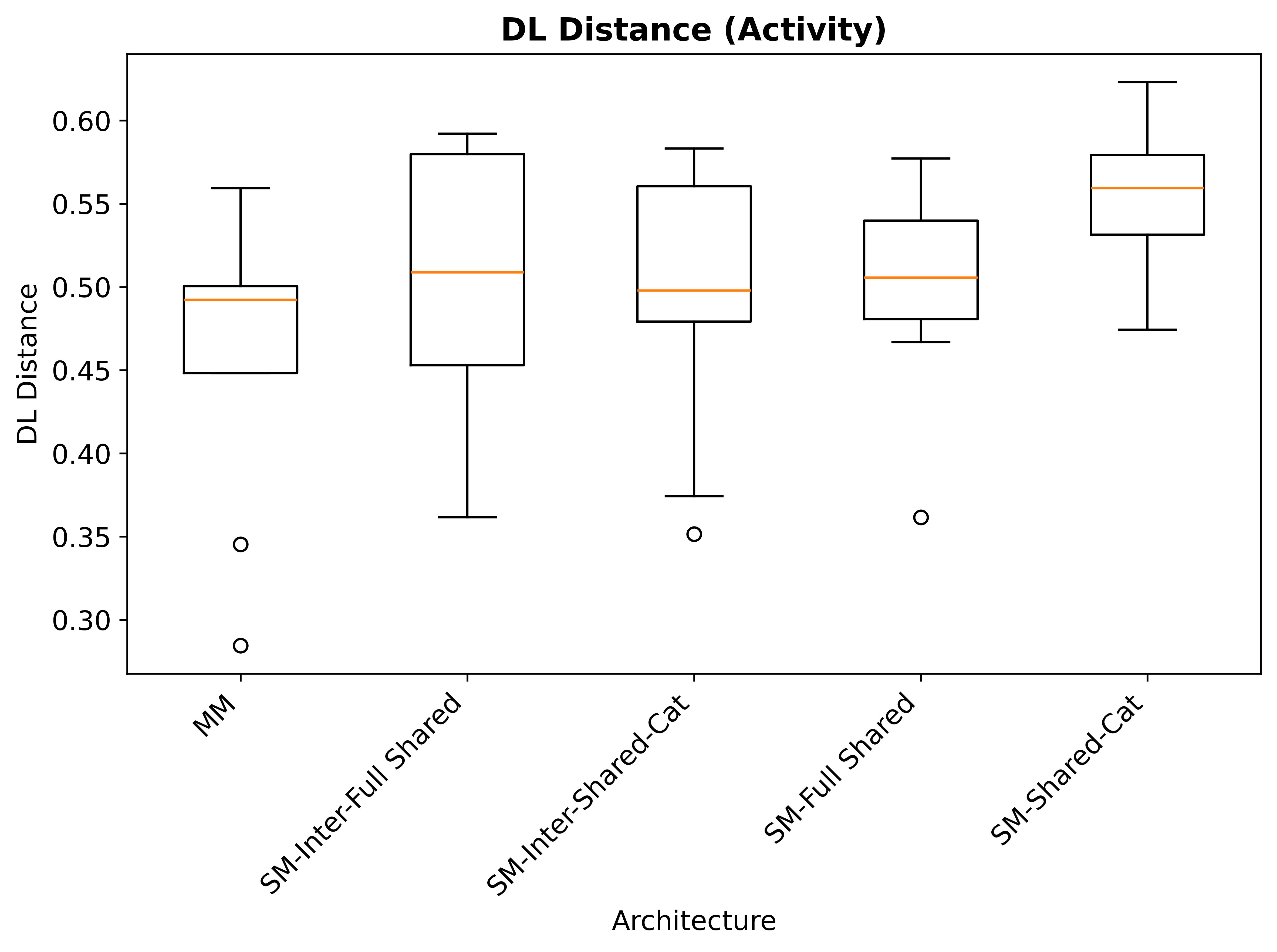}
        \caption{Distribution of Damerau-Levenshtein (DL) distance between MM and SM.}
        \label{fig:boxplot1}
    \end{subfigure}
    \hfill
    \begin{subfigure}[t]{0.48\linewidth}
        \centering
        \includegraphics[width=\linewidth]{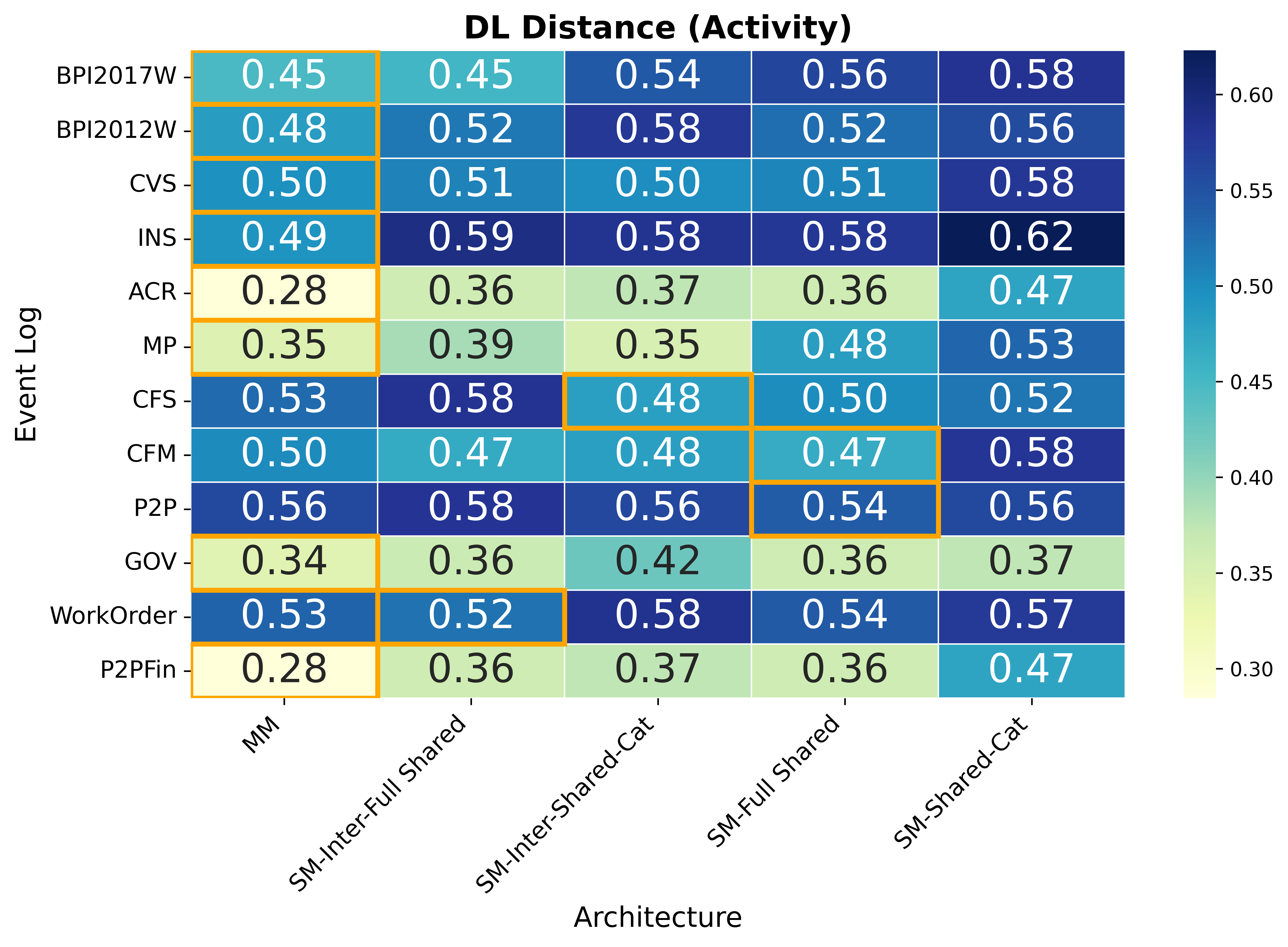}
        \caption{Heatmap showing DL distances between MM and SM across architectures.}
        \label{fig:heatmap1}
    \end{subfigure}
    \caption{DL distance (activity) comparison between Multi-Model (MM) and Single-Model (SM): (a) box plot, (b) heatmap. Results for \textbf{EQ1}.
}
 \label{EQ1}
\end{figure}

\begin{figure}[t]
    \centering
    \begin{subfigure}[t]{0.48\linewidth}
        \centering
        \includegraphics[width=\linewidth]{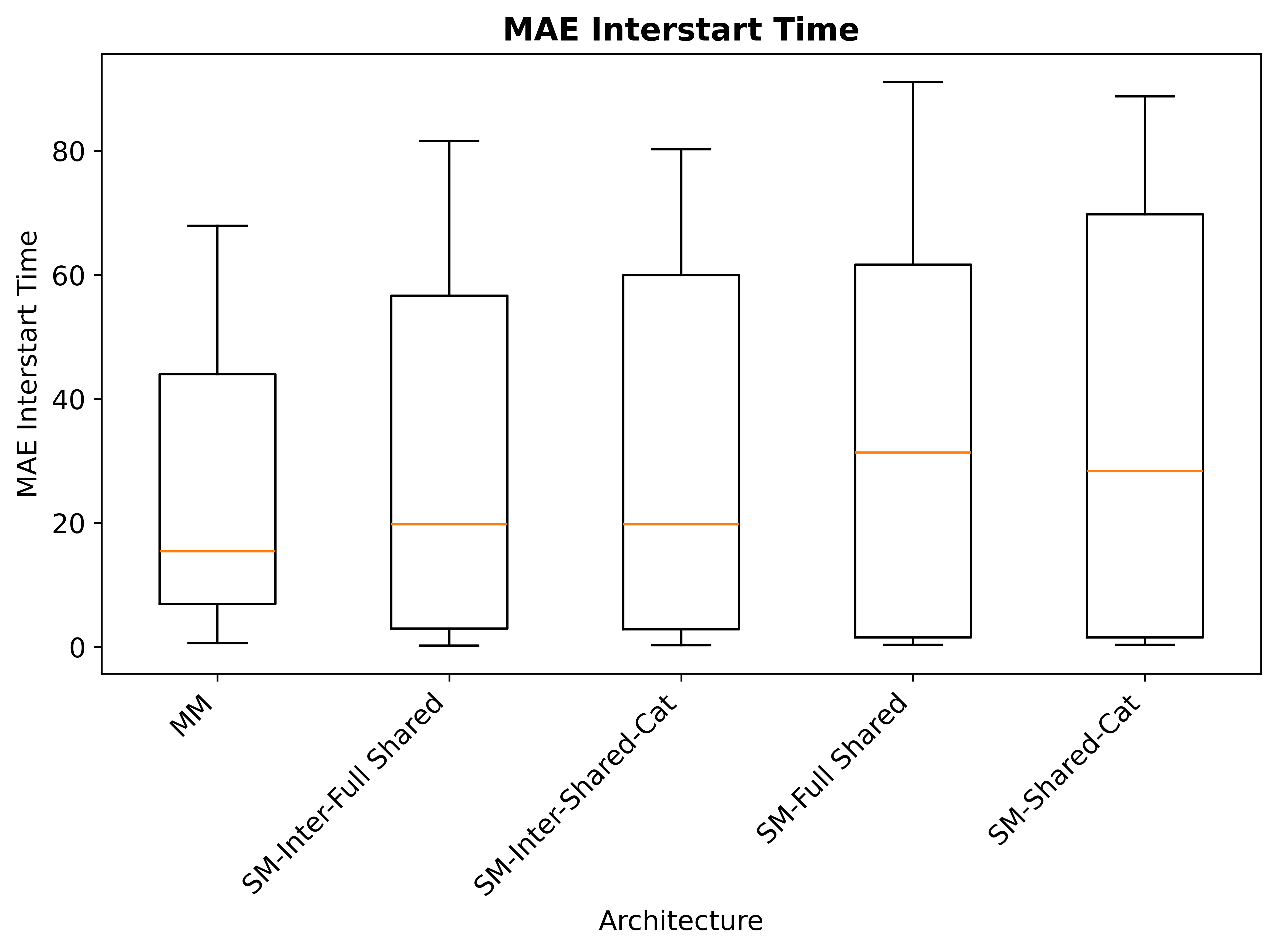}
        \caption{Box plot showing MAE of inter-start times for MM vs. SM.}
        \label{fig:boxplot2}
    \end{subfigure}
    \hfill
    \begin{subfigure}[t]{0.48\linewidth}
        \centering
        \includegraphics[width=\linewidth]{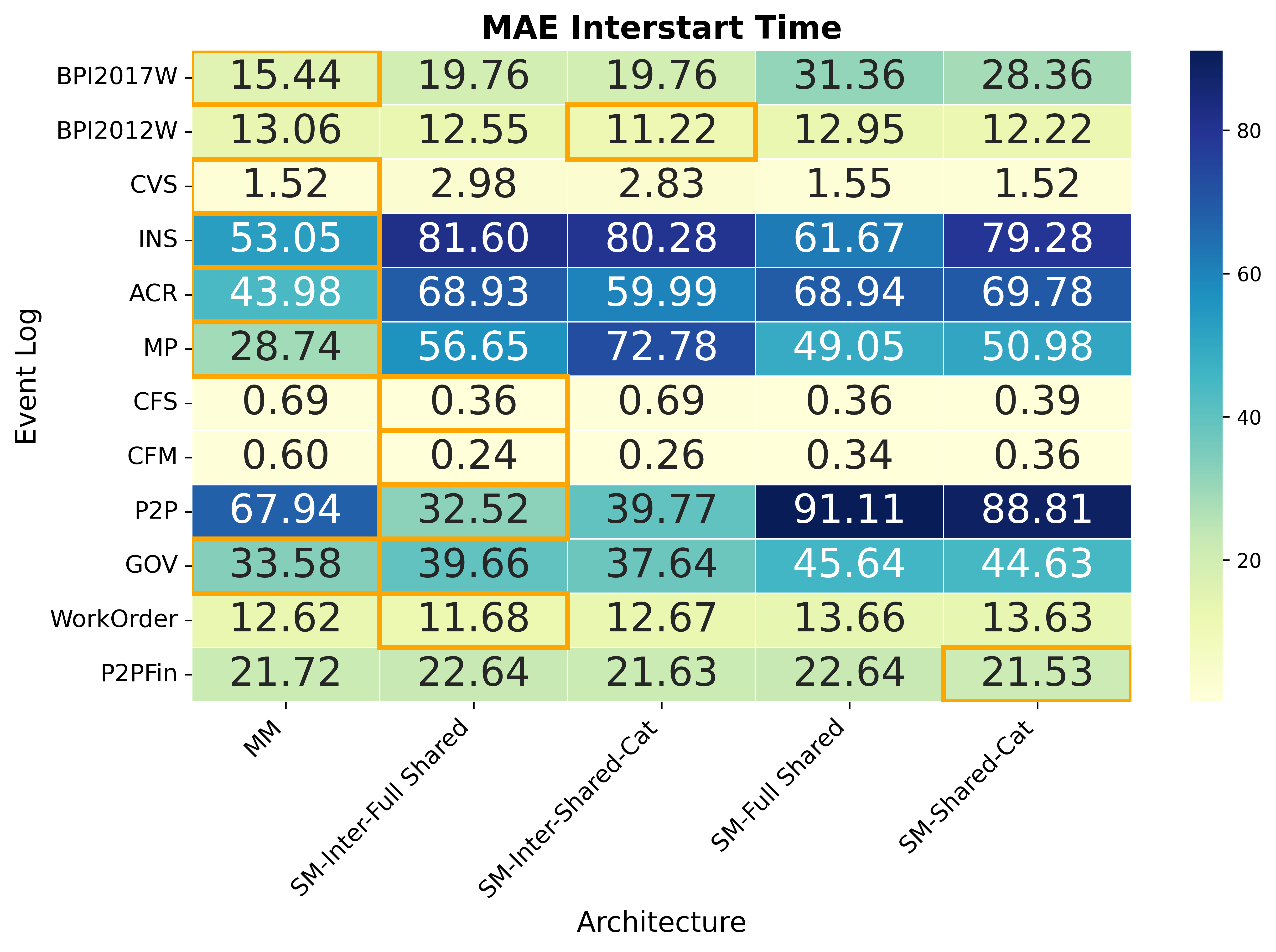}
        \caption{Heatmap of MAE in inter-start times between MM and SM architectures.}
        \label{fig:heatmap2}
    \end{subfigure}
    \caption{MAE in inter-start time predictions: Multi-Model (MM) vs. Single-Model (SM): (a) box plot, (b) heatmap. Results for \textbf{EQ2}.}
    \label{EQ2}
\end{figure}

\begin{figure}[htbp]
    \centering
    \begin{subfigure}[t]{0.48\linewidth}
        \centering
        \includegraphics[width=\linewidth]{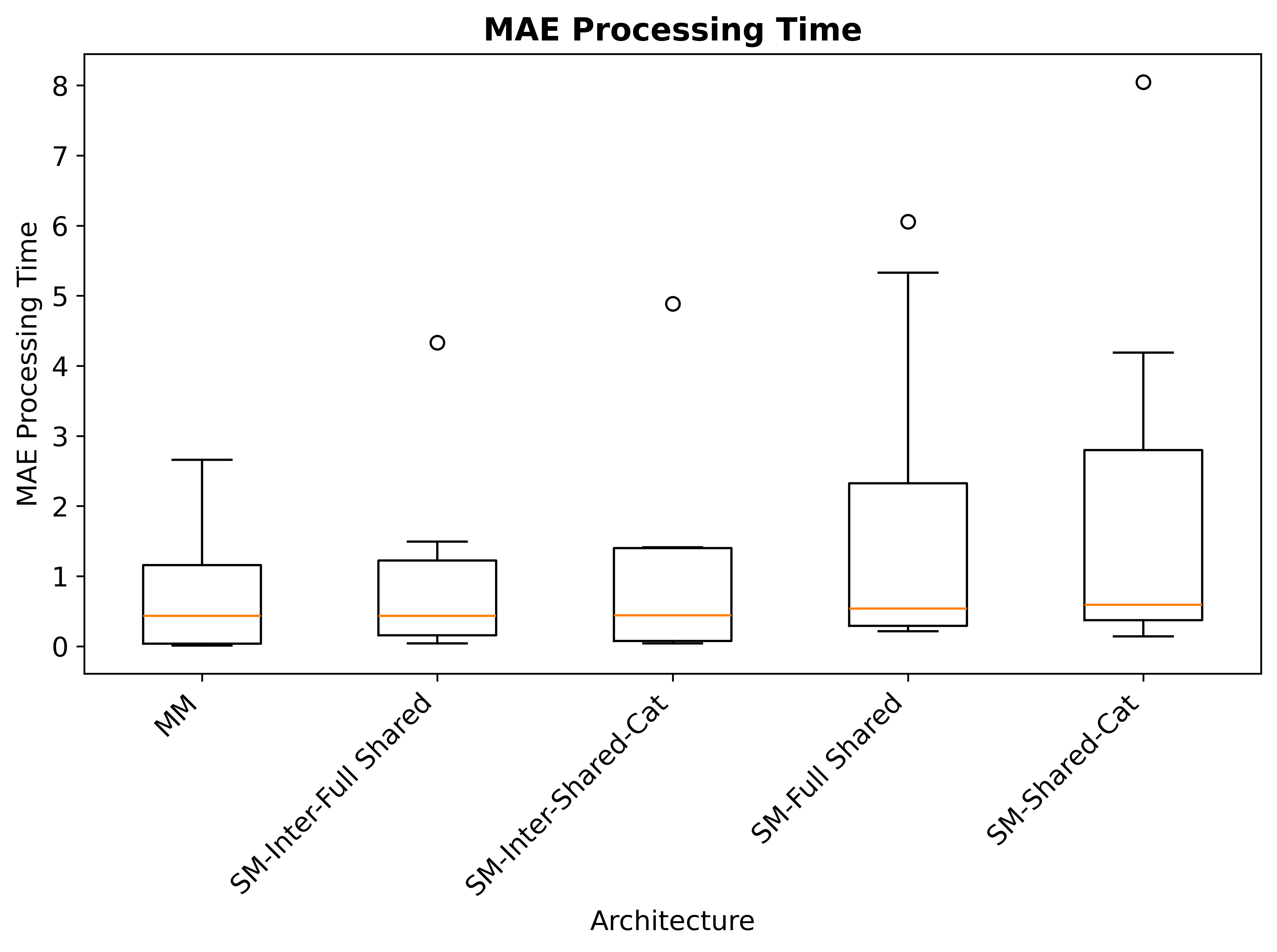}
        \caption{Box plot showing MAE of processing times for MM vs. SM.}
        \label{fig:boxplot3}
    \end{subfigure}
    \hfill
    \begin{subfigure}[t]{0.48\linewidth}
        \centering
        \includegraphics[width=\linewidth]{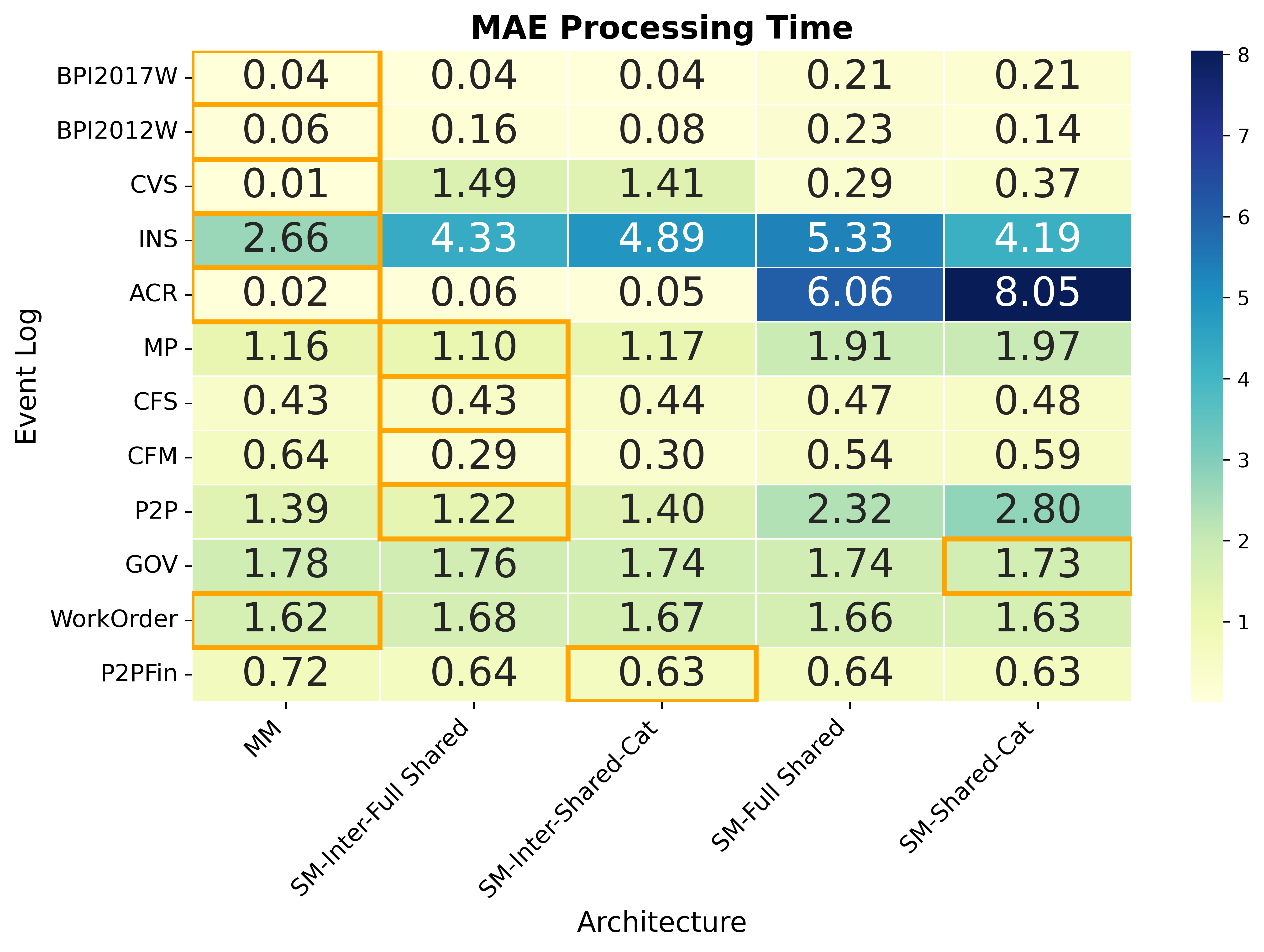}
        \caption{Heatmap of MAE in processing times between MM and SM architectures.}
        \label{fig:heatmap3}
    \end{subfigure}
    \caption{MAE in processing time predictions: Multi-Model (MM) vs. Single-Model (SM): (a) box plot, (b) heatmap. Results for \textbf{EQ3}.
}
    \label{EQ3}
\end{figure}

We evaluate the proposed architectures using three metrics: DL distance, inter-start time MAE, and processing time MAE. Summary results are shown in the box plots (Fig.~\ref{EQ1}a, Fig.~\ref{EQ2}a, and Fig.~\ref{EQ3}a), with each point representing a log. Detailed per-log results are provided in the corresponding heatmaps (Fig.~\ref{EQ1}b, Fig.~\ref{EQ2}b, and Fig.~\ref{EQ3}b), where bold-bordered cells mark the best-performing architecture per log. Lower values indicate better performance.

Regarding EQ1(a), we observe in Fig.~\ref{EQ1} that MM outperforms SM-based baselines on the control-flow metric. In the heatmap, we see that the outperformance of MM is most visible for event logs with fewer distinct activities (\textit{BPI2017W}, \textit{BPI2012W}, \textit{CVS}, \textit{INS}, and \textit{ACR}, which have less than 20 activities). When the number of activities is lower, the embeddings capture richer information about the sequential relations between activity pairs. The next-activity sampling method used in MM is then able to better exploit this information relative to the argmax next-activity sampling method used in the baselines. 

Regarding EQ1(b), Fig.~\ref{EQ2} box plot shows that the three techniques using resource contention features achieve slightly lower inter-start time MAE than the two baselines that rely only on intra-case features. 
The heatmap in Fig.~\ref{EQ2}b confirms this observation, particularly for event logs with fewer unique activities (\textit{BPI2017W}, \textit{CVS}, \textit{INS}, and \textit{ACR}).\footnote{An exception is \textit{BPI2012W}, where all the techniques exhibit similar performance.} For logs with larger number of distinct activities, we cannot make a clear conclusion. In the P2P log, the two baselines without resource contention features perform poorly, but in the CFS log, the use of resource contention features does not lead to lower inter-start time MAE and, in fact, MM has the worst performance among all techniques. 

Regarding EQ1(c), Fig.~\ref{EQ3} shows that resource contention features have minimal impact on processing time prediction. This is expected since processing time is determined by the activity's complexity rather than workload, and the resource contention features capture workload information. The  heatmap in Fig.~\ref{EQ3}b confirms this observation. It shows that MM outperforms the baselines in four of the five logs that have less than 20 distinct activities (\textit{BPI2012W}, \textit{CVS}, \textit{INS}, and \textit{ACR}), but there is no clear pattern in the remaining logs. The outperformance of MM on these four logs is attributable to the fact that it is able to better predict the next-activity, which in turn results in better predictions of the next activity's processing time. 

The above findings highlight that the use of embeddings has a positive effect in event logs with fewer distinct activities (\textit{BPI2017W}, \textit{BPI2012W}, \textit{CVS}, \textit{ACR}, \textit{INS}) but not in logs with 20+ activities. The embeddings we employ (taken from the method in~\cite{0001DR19}) are based on 3-grams, i.e. they look at which activities occur immediately before or after a given activity. To achieve better accuracy in logs with more distinct activities, we hypothesize that other types of embeddings are needed, e.g. hierarchical classification~\cite{SillaF11} or adaptive embeddings~\cite{SuccettiRP23}. 

Additionally, logs like \textit{INS} and \textit{MP} contain longer and more variable traces (mean case length $\approx$ 20 activities, CV of case length $>$ 65\%, with \textit{INS} having durations from 1 to nearly 600 days and a duration CV of 459\%), favoring MM’s modular design in learning decoupled control-flow and timing patterns. MM also excels on \textit{BPI2017W} despite a shorter mean case length ($\approx$ 7 activities), due to its extremely high duration variability (CV of duration 706\%). In contrast, \textit{CFS} and \textit{CFM} exhibit short (CV of case length $<$ 60\%) and homogeneous traces (mean duration $<$ 1 day, CV of duration 83\%), where low variability and limited concurrency make simpler SM architectures competitive.

To quantify the benefit of our proposed approach, we compared MM against the best-performing SM baseline for each metric. MM achieves a 6.89\% reduction in DL distance, indicating improved control-flow prediction accuracy. For inter-processing time prediction, MM reduces the mean absolute error (MAE) by 24.44\%, and for inter-start time prediction, by 18.35\%. These results confirm that MM not only improves predictive performance across all dimensions but does so with substantial error reductions over the most competitive baseline.

\subsection{Impact of Resource Availability Features}
We designed two experiments to evaluate different aspects of EQ2. The first experiment (EQ2.1) investigates whether the proposed resource availability features improve the performance of the MM approach using a synthetic event log where resources follow predefined (fixed) working calendars. The second experiment (EQ2.2) examines the impact of incorporating resource availability features on the predictive performance of the MM approach when applied to real-life event logs.

\subsubsection{Baselines}
For the evaluation of EQ2, we used the base MM approach without resource availability features as the baseline. This baseline represents the standard configuration of our approach, relying solely on inter and intra-case features without considering  resource availability $R_{avail}$. By comparing this baseline with the extended MM variant $MM_{avail}$ that incorporates resource availability features, we can directly assess the contribution of these features to predictive performance. This design allows us to isolate and quantify the effect of resource-related information on control flow, inter-start time, and processing time predictions across both synthetic and real-life event logs.

\subsubsection{Datasets}
To evaluate EQ2.1, we employed a simulation model of a loan application process as the baseline. From this model, we generated eight synthetic event logs, each corresponding to a different resource-availability scenario, as outlined in Table~\ref{scenarios}.

In the first scenario (a), all resources are continuously available, operating 24 hours a day, seven days a week. This results in a non-intermittent, dense, and homogeneous workload. In the second scenario (b), half of the resources share the same non-intermittent, dense schedule, working 24 hours a day from Monday to Sunday. The remaining half of the resources work 22 hours a day, Monday to Sunday, creating a heterogeneous workload. In the third scenario (c), all resources work 16 hours a day, Monday to Sunday, producing an intermittent, dense, and homogeneous workload. To create a heterogeneous workload in the fourth scenario (d), half of the resources work 16 hours a day, while the other half work 12 hours a day, Monday to Friday, resulting in an intermittent, dense, and heterogeneous calendar. To generate a sparse calendar, the resource workload should be less than 50 hours per week. In scenario (e), a resource works 8 hours a day, Monday to Friday, resulting in an intermittent, sparse, and homogeneous workload. In scenario (f), the workload is made intermittent by having half of the resources work 8 hours a day, while the other half work 4 hours a day, Monday to Friday. This produces an intermittent, sparse, and heterogeneous calendar. For a non-intermittent, sparse, and homogeneous workload, scenario (g) features a calendar where all resources work 8 hours a day on Mondays and Fridays only. In scenario (h), half of the resources work 6 hours a day on Tuesdays, Thursdays, and Saturdays, while the other half work 6 hours a day on Mondays, Wednesdays, and Fridays, creating an intermittent, sparse, and heterogeneous calendar. For each of these eight scenarios, we used the Apromore simulator to generate a synthetic event log consisting of 2,000 cases. The inter-arrival times between cases followed a Poisson distribution with a mean inter-arrival time of 20 minutes. Arrivals only occurred within the working hours defined by the corresponding resource calendars.

To evaluate EQ2.2, we used the same real-life event logs as in EQ1. The MM baseline without resource availability features was compared against the extended MM variant that incorporates these features $MM_{avail}$ to assess the impact on control flow, inter-start time, and processing time predictions.

\begin{table}[htbp]
\centering
\caption{Resource Availability Scenarios}
\footnotesize
\renewcommand{\arraystretch}{1.2}

\begin{tabular}{|c|>{\raggedright\arraybackslash}p{6.5cm}|>{\raggedright\arraybackslash}p{4cm}|}
\hline
\multicolumn{1}{|c|}{\textbf{Scenario}} &
\multicolumn{1}{c|}{\textbf{Resource Availability}} &
\multicolumn{1}{c|}{\textbf{Workload Type}} \\ \hline

a & 24 hours a day, 7 days a week
  & Non-intermittent, dense, homogeneous \\ \hline

b & Half resources: 24 hours a day, Monday to Sunday; half resources: 22 hours a day, Monday to Sunday
  & Non-intermittent, dense, heterogeneous \\ \hline

c & 16 hours a day, Monday to Sunday
  & Intermittent, dense, homogeneous \\ \hline

d & Half resources: 16 hours a day, Monday to Friday; half resources: 12 hours a day, Monday to Friday
  & Intermittent, dense, heterogeneous \\ \hline

e & 8 hours a day, Monday to Friday
  & Intermittent, sparse, homogeneous \\ \hline

f & Half resources: 8 hours a day, Monday to Friday; half resources: 4 hours a day, Monday to Friday
  & Intermittent, sparse, heterogeneous \\ \hline

g & 8 hours a day, Monday and Friday each week
  & Non-intermittent, sparse, homogeneous \\ \hline

h & Half resources work 6 hours a day on Tuesday, Thursday, and Saturday; half resources work 6 hours a day on Monday, Wednesday, and Friday
  & Non-intermittent, sparse, heterogeneous \\ \hline

\end{tabular}

\label{scenarios}
\end{table}

\subsubsection{Results and Interpretation}
Regarding EQ2.1, Table~\ref{EQ-2.1-results} presents the results for goodness measures (DL Distance, MAE Inter-start Time, and MAE Processing Time) for each resource availability scenario presented in Table~\ref{scenarios} for modified loan application process. These results provide key insights into how varying work shift scenarios for available resources affect the control-flow and temporal accuracy of case suffix prediction for MM and $MM_{avail}$.

The best performance in terms of DL Distance and MAE Inter-start Time is observed in Scenario (f), where resources are intermittently available, with half working 8 hours a day, Monday to Friday, and the other half working 4 hours a day, Monday to Friday. This scenario achieves the lowest values for DL Distance (0.33) and MAE Inter-start Time (3.81). Additionally, MAE Processing Time is also the best in this scenario (1.64). These results suggest that intermittent, sparse, and heterogeneous resource availability improves prediction accuracy. The model benefits from clear on-off-duty  cycles, reducing congestion and enabling more efficient task scheduling.


Conversely, Scenario (a), where resources are continuously available 24/7, yields higher DL Distance (0.43) and MAE Inter-start Time (3.97), along with a relatively high MAE Processing Time (1.73). This occurs because continuous availability leads to resource congestion and task overlap, which reduces the efficiency of load distribution and task management. The absence of structured work–rest cycles limits the model’s ability to learn meaningful patterns, and as a result, resource availability features have zero impact in this scenario because they do not vary over time and therefore provide no informative signal to the predictive model.

Scenarios with fixed working hours, like Scenario (c) with 16 hours per day, show intermediate performance. Specifically, Scenario (c) results in DL Distance (0.39), MAE Inter-start Time (3.87), and MAE Processing Time (1.68). While this schedule improves model performance compared to continuous availability, it still leads to higher errors compared to the more heterogeneous and sparse schedules, such as Scenario (f).

Scenario (e), with a sparser schedule of 8 hours per day, Monday to Friday, results in DL Distance (0.37) and MAE Inter-start Time (3.85), while maintaining lower processing times (1.68) compared to other scenarios. This indicates that a predictable, sparse resource schedule improves prediction accuracy, as the model can better anticipate resource availability, leading to more optimal scheduling.

The results indicate that intermittent and sparse resource schedules, particularly those with work hours distributed across blocks, improve prediction accuracy by 15.38\% for DL Distance, 3.05\% for inter-start time, and 3.53\% for processing time. Incorporating resource availability features therefore leads to measurable gains in the accuracy of predicting the next activity, inter-start times, and processing times.


\begin{table}[htbp]
\centering
\caption{EQ.2.1 - loan application process results}
\footnotesize
\renewcommand{\arraystretch}{1.5}  
\begin{tabular}{|c|c|c|c|c|c|c|c|c|}
\hline
\textbf{Scenerio}  & \multicolumn{2}{c|}{\shortstack{\textbf{DL} \\ \textbf{Distance}}} & \multicolumn{2}{c|}{\shortstack{\textbf{MAE} \\ \textbf{Inter-start Time}}} & \multicolumn{2}{c|}{\shortstack{\textbf{MAE} \\ \textbf{Processing Time}}} \\
\hline
 & {MM} & \textbf{$MM_{avail}$} & {MM} & \textbf{$MM_{avail}$} & {MM} & \textbf{$MM_{avail}$} \\
\hline
\textbf{a} &0.43 & \textbf{0.43} & 3.97 &\textbf{3.97}  &1.73  &\textbf{1.73}   \\
\hline
\textbf{b}&0.41 & \textbf{0.40} & 3.94 & \textbf{3.92} &1.73  & \textbf{1.72}  \\
\hline
\textbf{c} & 0.43 &\textbf{0.39}  & 3.91 &\textbf{3.87}  &1.71  &\textbf{1.68}   \\
\hline
\textbf{d} &0.40 & \textbf{0.37} & 3.90 &\textbf{3.84}  & 1.70 & \textbf{1.66} \\
\hline
\textbf{e} & 0.41 & \textbf{0.37 }& 3.92 & \textbf{3.85} & 1.72 & \textbf{1.68}  \\
\hline
\textbf{f} & 0.39& \textbf{0.33} &3.93  & \textbf{3.81} & 1.70 & \textbf{1.64}  \\
\hline
\textbf{g} & 0.42 & \textbf{0.34} & 3.92 & \textbf{3.84} & 1.71 &\textbf{1.66}   \\
\hline
\textbf{h} &0.43 & \textbf{0.37} & 3.93 & \textbf{3.86} & 1.72 &\textbf{1.67}  \\
\hline

\end{tabular}

\label{EQ-2.1-results}
\end{table}



Regarding EQ2.2, Table~\ref{tab:MM VS MMavail} presents a comparison between MM and $MM_{avail}$ for predicting case suffixes in an ongoing business process. The empirical results demonstrate that $MM_{avail}$ consistently outperforms the baseline across all three evaluation metrics: DL Distance, MAE Inter-start Time, and MAE Processing Time. Notably, $MM_{avail}$ reduces the DL Distance, for instance, from 0.48 to 0.39 in the BPI2012W dataset, representing a substantial improvement of 18.75\%. This significant reduction indicates a refined capability to capture control-flow fluctuations. This improvement is especially noticeable in datasets with variable resource schedules, such as INS, MP, and P2P. $MM_{avail}$ is able to account for resource availability gaps by incorporating resource availability features, providing a more accurate prediction. The table also highlights that these real-life logs tend to have less dense(Den.) and more sparse resource calendars, with the highest incidence of intermittent(Int.) schedules. Moreover, the availability features such as \textit{Time Until Next On-Duty} and \textit{Time Since Last Off-Duty} enable the $MM_{avail}$ model to better understand resource on-off-duty  patterns. Consequently, this mitigates sequence alignment errors, yielding superior chronological precision for the control flow of ongoing cases.


In terms of MAE Inter-start Time, $MM_{avail}$ consistently outperforms MM, with significant improvements observed in datasets such as INS (from 53.05 to 45.04) and ACR (from 43.98 to 36.02). These datasets feature highly intermittent and sparse resource availability calendars, where accurately predicting the time between task completions requires an understanding of resource on-off-duty  cycles. The resource availability features of $MM_{avail}$ enable it to effectively manage these intermittent and sparse schedules, leading to more precise predictions of when the next task should begin. In contrast, MM struggles with such schedules, as it does not account for resource availability, resulting in less accurate predictions of inter-start times for the predicted case suffixes.

For the MAE Processing Time, both MM and $MM_{avail}$ perform similarly on certain logs, such as BPI2017W and CVS. This similarity arises because processing time is less influenced by workload factors like resource availability. Instead, it is primarily determined by the nature of the activity itself. However, $MM_{avail}$ shows clear advantages on datasets like Work Order, INS, and P2PFin, where the coefficient of variation (CV) for case length varies between 48\% and 74\%, and the CV for case duration varies between 149\% and 459\%. On other logs, the improvement remains either consistent or improved with respect to the MM approach. Hence, the inclusion of availability features enhances the model's ability to predict task durations, factoring in the on-off-duty  cycles of resources. $MM_{avail}$ demonstrates better performance, particularly on event logs with sparse and highly intermittent resource on-off-duty  schedules, where the availability features lead to more accurate predictions compared to the MM approach.

In conclusion, $MM_{avail}$ demonstrates superior performance compared to the $MM$ approach for all three metrics. This improvement is fundamentally linked to the accuracy of start-time predictions, which governs the correctness of the DL distance. Accurate start times are essential because the model processes activity instances based on their sorted chronological order. For instance, consider two overlapping activities: one starting at 10 and another at 12. If the start time of the second activity is underestimated (e.g., predicted as 9), the sorting logic inevitably inverts their sequence. Such disordering negatively impacts the sweep-line based technique used by the model. The $MM_{avail}$ method addresses this by leveraging resource availability features to refine start-time estimates, thereby preserving the correct sequence of activities and significantly improving both control flow accuracy and processing time prediction.


\begin{table}[htbp]
\centering
\caption{EQ.2.2 - Comparison of MM and $MM_{avail}$ Approaches}
\footnotesize
\renewcommand{\arraystretch}{1.5}  
\begin{tabular}{|c|c|c|c|c|c|c|c|c|}
\hline
\textbf{Dataset} &\textbf{Den.} &\textbf{Int.} & \multicolumn{2}{c|}{\shortstack{\textbf{DL} \\ \textbf{Distance}}} & \multicolumn{2}{c|}{\shortstack{\textbf{MAE} \\ \textbf{Inter-start Time}}} & \multicolumn{2}{c|}{\shortstack{\textbf{MAE} \\ \textbf{Processing Time}}} \\
\hline
&& & {MM} & \textbf{$MM_{avail}$} & {MM} & \textbf{$MM_{avail}$} & {MM} & \textbf{$MM_{avail}$} \\
\hline
\textbf{BPI2017W} &33.04\% & 0.36& 0.45& \textbf{0.43} & 15.44 & \textbf{14.01} & 0.04 & \textbf{0.04}  \\
\hline
\textbf{BPI2012W}&26.36\%& 0.43& 0.48 & \textbf{0.39} & 13.06 & \textbf{11.05} & 0.06 & \textbf{0.05} \\
\hline
\textbf{CVS} &39.31\%&0.27& 0.50 & \textbf{0.50} & 1.52 & \textbf{1.43} & 0.01 & \textbf{0.01} \\
\hline
\textbf{INS} &10.82\%&0.36& 0.49 & \textbf{0.44 }& 53.05 & \textbf{45.04} & 2.66 & \textbf{2.43}  \\
\hline
\textbf{ACR} &3.04\%&0.23& 0.28 & \textbf{0.26} & 43.98 & \textbf{36.03} & 0.02 & \textbf{0.02} \\
\hline
\textbf{MP} &28.27\%&0.45& 0.35 & \textbf{0.27} & 28.74 & \textbf{24.05} & 1.16 & \textbf{1.12} \\
\hline
\textbf{CFS} &43.41\%&0.25& 0.53 & \textbf{0.53} & 0.69 & \textbf{0.63} & 0.43 & \textbf{0.39} \\
\hline
\textbf{CFM} &34.44\%&0.24& 0.50 & \textbf{0.48} & 0.60 & \textbf{0.55} & 0.64 & \textbf{0.61} \\
\hline
\textbf{P2P} &32.01\%&0.31& 0.56 & \textbf{0.48} & 67.94 & \textbf{64.82} & 1.39 & \textbf{1.37} \\
\hline
\textbf{GOV} &32.96\%&0.41& 0.34 & \textbf{0.28} & 33.58 & \textbf{27.98}& 1.78 & \textbf{1.76} \\
\hline
\textbf{Work Order} &21.13\%&0.48& 0.53 & \textbf{0.51} & 12.62 & \textbf{9.76} & 1.62 & \textbf{1.58} \\
\hline
\textbf{P2PFin} &11.31\%&0.29& 0.28 & \textbf{0.27}  & 21.72 & \textbf{18.43} & 0.72 & \textbf{0.67} \\
\hline

\end{tabular}

\label{tab:MM VS MMavail}
\end{table}

\section{Conclusion}\label{conclusion}
This paper presents a method to predict the sequence of remaining activity instances for each ongoing case of a business process, along with the inter-start time and processing time of each activity instance. 
The proposed method predicts the suffixes of all ongoing cases in lockstep, using a sweep line-based approach. At each cut-off point, starting from  the time of the most recently started activity instance, we estimate the label of the next activity instance in the same case, and the (inter-)start time and processing time of this activity instance.
After each such next-activity prediction step, we recalculate a set of features related to resource contention and resource availability. Resource contention and availability are two key sources of waiting time in business processes. Accordingly, the proposed method builds on the hypothesis that the inclusion of such features is likely to enhance the accuracy of inter-start time predictions.

The proposed method predicts case suffixes using three separate models: one for the next activity prediction, another for inter-start time prediction, and a third for processing time prediction. This multi-model approach enables us to optimize each model separately.

The experimental evaluation shows that the proposed multi-model (MM) method achieves higher accuracy relative to existing baselines, which predict the activity and timestamps via a single model (SM approaches) and which do not incorporate resource contention and resource availability features. In particular, we observed an improvement in the control-flow accuracy metrics, attributable to the fact that we exploited the modularity of the proposed approach to incorporate an optimized method for next-activity sampling. We also observed improvements in inter-start time prediction attributable to the use of resource contention features in the proposed sweep-line method. Furthermore, we observed that when the resource availability features are integrated within the MM approach, this information significantly improved the performance of MM’s variant $MM_{avail}$ for case suffix prediction. This improvement was particularly evident with respect to control flow, inter-start time, and processing time prediction. In comparison to the previous MM approach, $MM_{avail}$ demonstrated improved performance, highlighting the effectiveness of incorporating resource availability features into the predictive model for predicting case suffix of an ongoing case in a business process.

\section{Future Work}\label{FW}

In this paper, we developed predictive models that leverage features capturing two primary sources of waiting times in business processes: resource contention and resource availability. While these factors account for a significant portion of waiting times in business processes, previous studies by Ali et al.~\cite{ali_data-driven_2024} and Lashkevich et al.~\cite{LashkevichMCSD23} have shown that waiting times also arise from other sources, including batching effects, prioritization policies, and external factors beyond the scope of the process.

A promising direction for future work is to extend the proposed approach by incorporating features that explicitly model these additional sources of waiting time. For example, future approaches could capture batching behaviour by representing batch formation rules and batch sizes, account for prioritization by modelling case-level or activity-level priority schemes, and include external contextual information (e.g., calendar effects or unexpected disruptions) to better explain externally induced delays. Integrating these factors is expected to further improve the accuracy and robustness of waiting time predictions.

In addition, we plan to apply the proposed approach to practical capacity planning and scheduling use cases. In particular, the learned models could be used by process managers to evaluate and optimize alternative resource availability schedules given the current state of a running process. For instance, in scenarios where a large number of cases are pending and additional resources (e.g., workers) can be temporarily reassigned, the proposed approach could be used to answer ``what-if'' questions such as: Given the current workload and the availability of ten additional workers, what resource schedules would best ensure that ongoing cases meet their deadlines? Addressing such decision-support scenarios represents an important step toward operationalizing predictive process monitoring models in real-world settings.



\smallskip\noindent\textbf{Reproducibility.} Source code and  Supplementary Material: \\ \href{https://github.com/AwaisAli37405/How-Will-My-Business-Process-Unfold-Predicting-Case-Suffixes-With-Start-and-End-Timestamps.git}{https://shorturl.at/wcu1P}, 
\href{https://figshare.com/s/cd6a11b081d80e3a9a28}{https://shorturl.at/PM3aV}

\medskip\noindent\textbf{Acknowledgments.} 
Work funded by the European Research Council (PIX Project) and the Estonian Research Council (PRG1226).

\bibliographystyle{ws-ijcis}
\bibliography{sample}

\end{document}